\documentclass[fleqn,10pt]{wlscirep}
\usepackage[utf8]{inputenc}
\usepackage[T1]{fontenc}
\usepackage{multirow}
\RequirePackage{graphicx}
\RequirePackage{fancyhdr}
\RequirePackage{bookmark}
\RequirePackage{caption} 
\usepackage{arydshln}
\usepackage{lineno}
\RequirePackage{amsthm}

\usepackage[nameinlink]{cleveref}

\crefname{figure}{Fig.}{Figs.}
\crefformat{equation}{Eq.~#2(#1)#3}
\crefformat{section}{Section~#2#1#3}
\AtBeginDocument{%
	\let\citet\cite
}

\usepackage[labelfont=bf]{caption}
\usepackage{babel}

\usepackage{xcolor}
\definecolor{docnotelinkcolor}{rgb}{0,0,0.5}
\usepackage{hyperref}
\hypersetup{colorlinks = true, allcolors = blue}

\theoremstyle{thmstyleone}

\title{\fontsize{18}{23}\selectfont Revealing Imaging Characteristics of Thermal Wakes from Underwater Vehicles Without Cooling-Water Discharge}

\author[1]{Guoyi Zhang}

\author[1]{Yanjin Du}

\author[1]{Guangsheng Xu}

\author[1]{Zhengyao Zhao}

\author[1]{Siyang Chen}
	
\author[1]{Jingjing Gao}

\author[1]{Yuezheng Zhou}

\author[1]{Yiyang Wu}

\author[1]{Bin Wu}

\author[1]{Tongsu Zhang}

\author[1]{Lejia Ye}

\author[2]{Donghe Wang}

\author[3]{Jie Wang}

\author[4]{Kaiye Hu}

\author[5]{Zhipeng Ren}

\author[1,*]{Han Wang}

\author[6,*]{Boo Cheong Khoo}

\author[1,*]{Xiangpeng Xu}

\author[1,*]{Xiaohu Zhang}

\affil[1]{School of Aeronautics and Astronautics, Sun Yat-sen University,  Shenzhen, 518107, Guangdong, China}

\affil[2]{Changchun Institute of Optics, Fine Mechanics and Physics, Chinese Academy of Sciences, Changchun 130033, Jilin, China}

\affil[3]{Institute of Machinery Manufacturing Technology, China Academy of Engineering Physics, Mianyang, 621900, China}

\affil[4]{College of Shipbuilding Engineering, Harbin Engineering University, Harbin, 150006, Heilongjiang, China}

\affil[5]{School of Energy Science and Engineering, Harbin Institute of Technology, Harbin, 150001, Heilongjiang, China}

\affil[6]{Mechanical Engineering Department, National University of Singapore, 117575, Singapore}

\affil[*]{Corresponding. zhanggy57@mail2.sysu.edu.cn, duyj36@mail2.sysu.edu.cn, duyj36@mail2.sysu.edu.cn, xugsh6@mail2.sysu.edu.cn, zhaozhy65@mail2.sysu.edu.cn, chensy253@mail2.sysu.edu.cn, gaojj27@mail2.sysu.edu.cn, zhouyzh28@mail2.sysu.edu.cn, wuyy239@mail2.sysu.edu.cn, wubin58@mail2.sysu.edu.cn, zhangts3@mail2.sysu.edu.cn, yelj8@mail2.sysu.edu.cn, 15104425162@163.com, wangj688@alumni.sysu.edu.cn, hukaiye@hrbeu.edu.cn, renzhipeng@hit.edu.cn, wangh737@mail2.sysu.edu.cn, mpekbc@nus.edu.sg, xuxp39@mail.sysu.edu.cn, zhangxiaohu@mail.sysu.edu.cn}

\begin{abstract}
Small uncrewed underwater vehicles are spreading rapidly and threaten maritime security, yet they evade the sonar and magnetic sensors built to track larger submarines. Infrared thermal wakes, generated by hydrodynamic displacement that modulates the cold-skin layer, offer a non-acoustic, non-magnetic alternative even without cooling-water discharge. However, existing studies rely on idealized simulations that fail to capture complex real-world backgrounds, offering limited guidance for target-background discrimination. Here we establish AVP-IR-Wake, the first large-scale, real-world, physically diverse benchmark of 161,485 frames collected over three years. We show that thermal wakes are extended structures composed of discrete vortices, which fundamentally violate the compactness assumption of conventional targets. More importantly, we reveal a fundamental local indistinguishability: individual wake vortices are statistically identical to wind-induced background vortices in the temporal, frequency, and sparse domains, sharing identical low-frequency information and long-range correlations. This directly challenges the prevailing assumptions of compact, transient, and high-frequency targets in conventional detection paradigms. Yet, we establish a global morphological separability: although locally inseparable, the wake as a whole is organized by the vehicle trajectory, whereas background clutter is organized by wind-driven thermal fronts. This intrinsic imaging property explains why representative detectors fail systematically and demonstrates that reliable detection must transition from local feature extraction to global trajectory-constrained physical modeling. Because the cold-skin layer varies globally and seasonally, forecasting wake observability will require integrating physical oceanography and atmospheric science, linking underwater surveillance directly to the changing ocean surface.
\end{abstract}

\begin{document}

\flushbottom
\maketitle 
\section*{Introduction}
The ocean underpins global ecosystems, trade and maritime security, yet these interests face growing challenges from the rapid proliferation of small uncrewed underwater vehicles \cite{dfm2025natosea,ukgov2026reef}. Inexpensive, quiet and increasingly widespread, these platforms can threaten ports and critical subsea infrastructure while evading sonar and magnetic sensors primarily designed to detect larger submarines \cite{laursen2026intruder,rossiter2025cable}. In 2026, the U.S. Coast Guard acknowledged that conventional approaches cannot reliably detect, track or mitigate these vehicles \cite{USCG2026CUUV}. Their acoustic signatures can be masked by reverberation in shallow water and at low speeds \cite{railey2021passive}, while their weak magnetic signatures can be obscured by geomagnetic and environmental noise \cite{wu2022research,liu2023theories}. These limitations have prompted increased investment in underwater surveillance, from the UK–US REEF programme to NATO–EU initiatives addressing undersea infrastructure security \cite{ukgov2026reef,dfm2025natosea}. Together, these developments underscore the need for complementary sensing approaches that can reveal underwater vehicles independently of their acoustic and magnetic signatures.

Infrared thermal wake detection offers a non-acoustic, non-magnetic approach to underwater surveillance. Early studies attributed these signatures to cooling-water discharge \cite{wang2023numerical}, but propeller-induced mixing accelerates energy dissipation \cite{han2026hull}, modern propulsion systems reduce discharge temperature and volume \cite{gao2022influence}, and many small uncrewed underwater vehicles release no cooling water at all \cite{ijspeert2026swimming,scirobotics.adx9939,liu2024maneuverable}. Recent fluid-physics studies have identified an alternative mechanism \cite{luo2023thermal}: as a vehicle moves, its volumetric displacement transports warmer subsurface water towards the surface \cite{voropayev2009surface}, perturbing the cold-skin layer, a sub-millimetre inversion layer formed by air-sea heat exchange \cite{he2024common}. This process can generate infrared thermal wakes without cooling-water discharge \cite{luo2024experimental}, with morphology and physical characteristics that differ substantially from those of discharge-induced wakes \cite{wang2023numerical,wang2023surface,wang2021numerical}. However, because the cold-skin layer is maintained by natural thermodynamic conditions that are difficult to reproduce in indoor wave tanks, displacement-induced wakes remain challenging to investigate experimentally. Although controlled experiments have shown that surface wake signatures depend strongly on navigation speed, depth and environmental conditions \cite{gao2025airborne}, and numerical studies have modelled wake temperature fields \cite{wang2026surface,luo2024reconstruction} and simulated their infrared imaging \cite{ma2015monte}, these approaches cannot fully capture real-world environmental variability. In particular, the environmental conditions governing cold-skin dynamics and wake evolution are difficult to measure and model, leaving the imaging characteristics of displacement-induced thermal wakes under realistic and varying conditions insufficiently understood (Supplementary Figure \ref{fig:Background}).

However, from a detection perspective, merely characterizing the physical properties of a thermal wake is not sufficient. The key challenge is to distinguish the wake from complex natural backgrounds \cite{10606508,xiong2026sam2}, whose thermal structures may resemble those of the target. Thus, even if the physical conditions governing wake formation are understood, it remains uncertain whether the resulting infrared signatures can be reliably distinguished from environmental clutter. Absolute signal intensity or isolated thermal anomalies may be insufficient for this purpose because target signatures can overlap with naturally occurring environmental structures \cite{wang2026eaa}. Furthermore, controlled experiments and simulated datasets often involve simplified scenarios \cite{gao2025airborne} that do not adequately represent the diversity of real-world backgrounds, including wind-driven thermal fronts. Even under ideal conditions, the infrared imaging process itself introduces significant discrepancies: due to system nonlinearities, limited camera sensitivity, atmospheric transmission loss, and atmospheric turbulence \cite{wang2023revelation}, the morphological appearance of wake targets in real infrared images can deviate substantially from the idealized physical temperature fields. Targets in controlled experiments are also often constrained to fixed trajectories \cite{gao2025airborne}, limiting their appearance to specific image regions and potentially introducing positional bias. Deep learning models trained on such data may exploit these spatial correlations rather than learn features intrinsic to the thermal wake \cite{10547427}. Compounding these challenges, acquiring large-scale, real-world detection data is exceptionally difficult. Field trials are resource-intensive and prohibitively expensive, while the sensitive nature of underwater vehicle surveillance often restricts data sharing. Moreover, previous data collection efforts were predominantly physics-driven \cite{gao2025analysis} rather than tailored for computer vision, resulting in a critical scarcity of scene-diverse data required to train robust detectors. These limitations highlight the need to characterize thermal wakes in uncontrolled, physically diverse environments, where their observable properties can be examined alongside the background structures that complicate detection.

However, it remains unclear whether the assumptions underlying conventional detection paradigms hold in complex real-world environments. Conventional infrared dim-target detection typically relies on three assumptions. First, targets are spatially compact \cite{zhang2025novel}. Second, targets are temporally transient or sparse relative to the background \cite{11156113}. Third, targets occupy a distinct spectral band, typically at higher spatial frequencies than the background \cite{11373245}. These assumptions have proven effective for detecting small, rigid, hot objects such as aircraft, ships and satellites \cite{11146868}. Thermal wakes, however, differ fundamentally from such targets. Generated by fluid motion rather than a compact heat source, they can extend over large areas, persist for minutes and comprise multiple discrete vortices (hereafter referred to in an imaging sense as localized thermal anomalies). It therefore remains unclear whether conventional detection assumptions hold for thermal wakes and which imaging properties, if any, can distinguish them from natural background clutter.

To address this gap, we establish AVP-IR-Wake, the first large-scale, real-world benchmark comprising 161,485 infrared frames collected over three years across multiple lakes and vehicle types. We develop a physically grounded annotation protocol that integrates trajectory consistency, hysteresis effects, CFD-assisted simulation and expert cross-validation to localize weak and ambiguous wakes. Using this dataset, we find that individual wake vortices are statistically indistinguishable from wind-induced background vortices in the temporal, frequency and sparse domains. This local indistinguishability challenges the assumptions underlying conventional detection paradigms. Yet, at the global scale, the wake retains a distinct morphological organization constrained by the vehicle trajectory, whereas background clutter is organized by wind-driven thermal fronts. Evaluations of representative detectors further show that conventional detection priors, including compactness, high-frequency content, sparsity and temporal profiles, are insufficient to reliably identify these wakes. Together, these findings point to the need for detection approaches that move beyond local feature extraction towards global, trajectory-constrained physical modeling. Because wake observability depends on the coupled ocean-atmosphere boundary layer, which varies across regions and seasons, understanding thermal wakes may also provide insights into how environmental conditions shape underwater vehicle observability and stealth.
\section*{Results}
\bigbreak
\noindent
\textbf{Statistical Characterization of AVP-IR-Wake}

\begin{figure*}[!t]
	\centering
	\includegraphics[width=\linewidth]{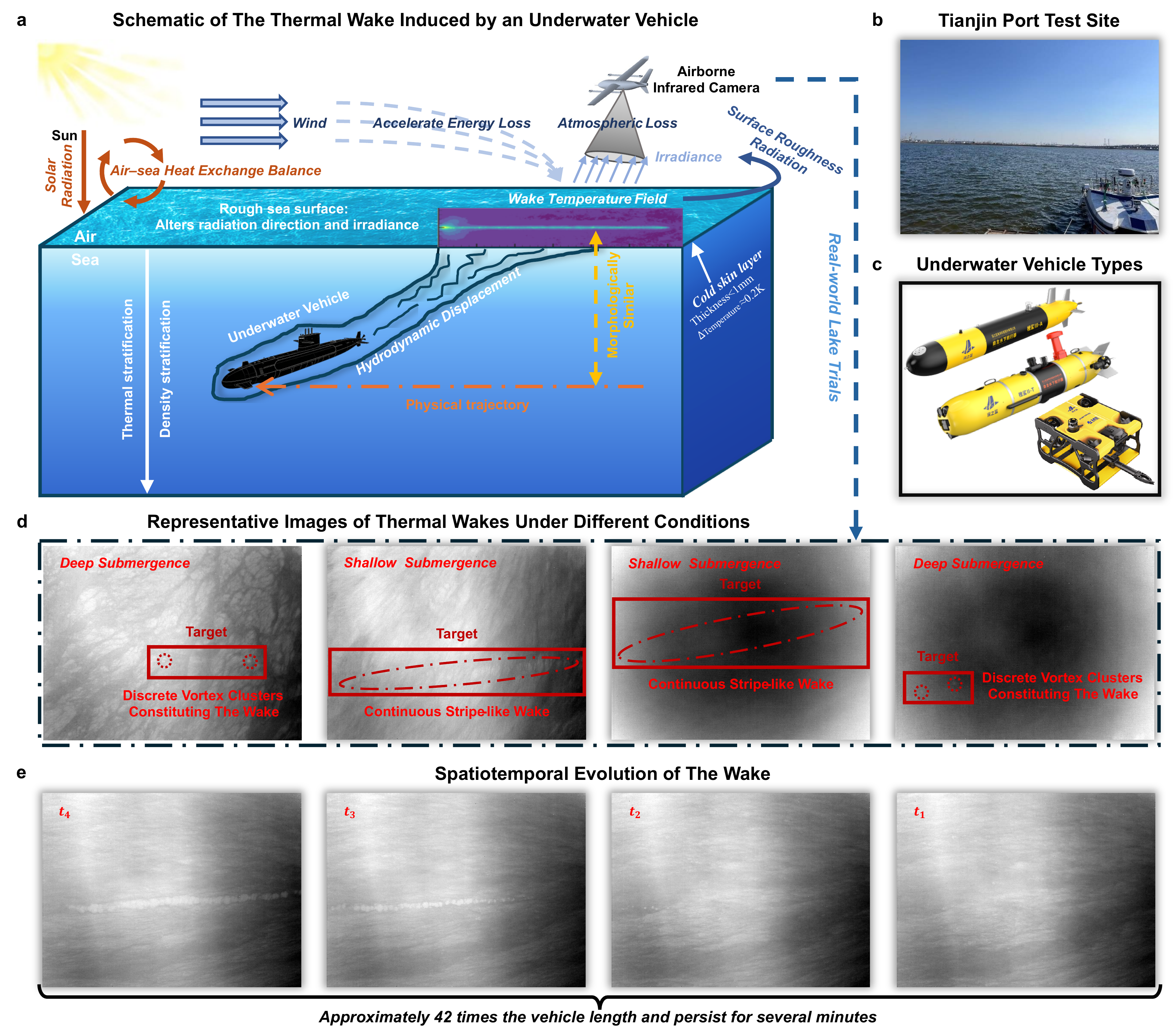}
	\caption{\textbf{Thermal wakes as infrared visual targets.} \textbf{a}, Schematic of wake formation. The vehicle’s hydrodynamic displacement transports warmer subsurface water upward, modulating the cold-skin layer and producing a surface thermal signature even without cooling-water discharge. 
		\textbf{b} and \textbf{c}, Data were mainly collected at Tianjin Port using both AUVs and ROVs. 
		\textbf{d}, Representative images of thermal wakes under different conditions. Deep submergence produces discrete vortex clusters, while shallow submergence produces continuous stripe-like wakes. These images show that the same target can appear in fundamentally different forms depending on operating conditions.
		\textbf{e}, The wake extends approximately 42 times the vehicle length and persists for several minutes, far exceeding the scale of the vehicle itself.}
	\label{fig:Compare}   
\end{figure*}
\begin{figure*}[!t]
	\centering
	\includegraphics[width=\linewidth]{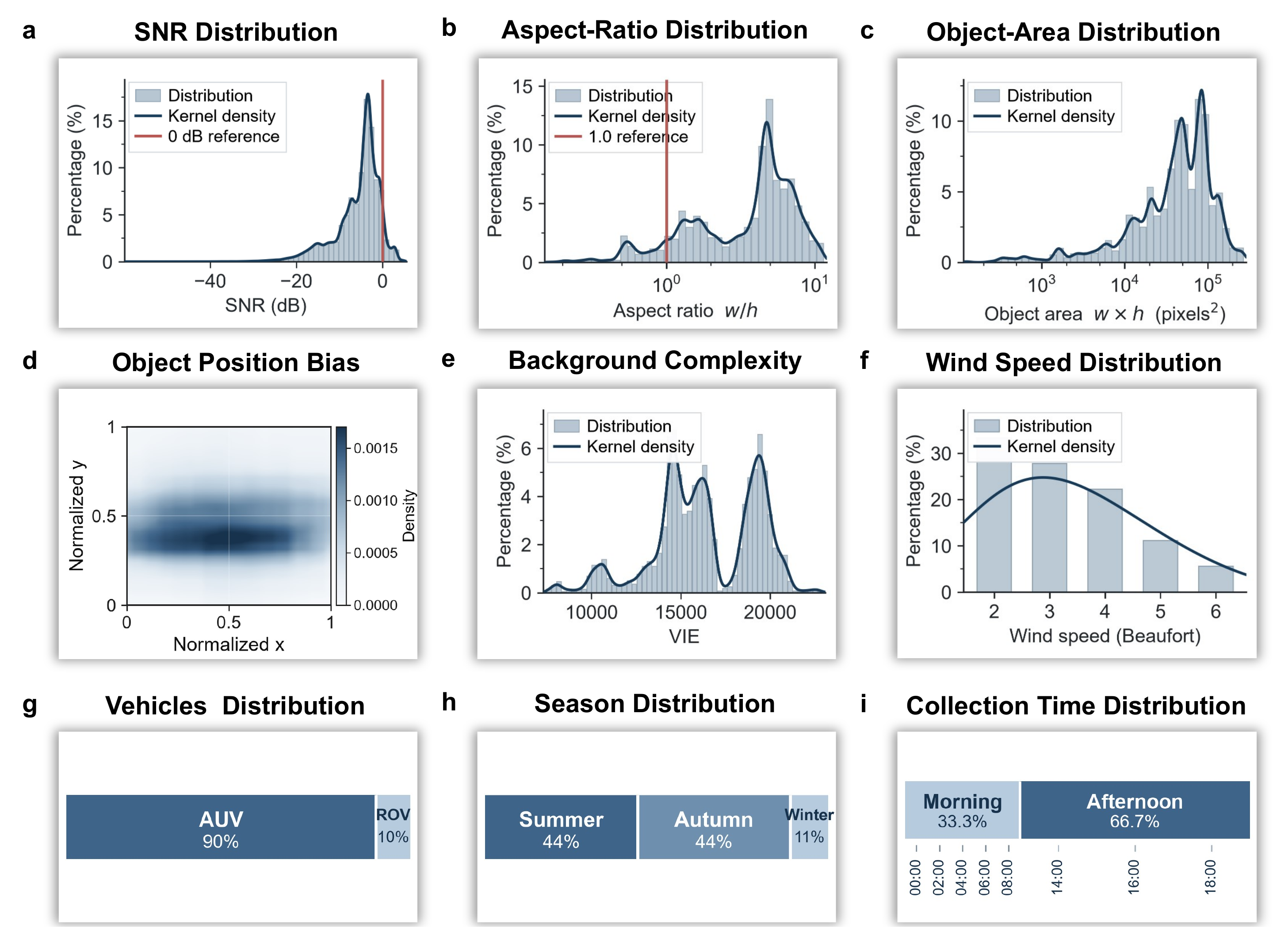}
	\caption{\textbf{Statistical characterization of AVP-IR-Wake.} \textbf{a}, SNR distribution. 
		\textbf{b}, aspect-ratio  distribution.
		\textbf{c}, object-area distribution.
		\textbf{d}, object position bias.
		\textbf{e}, background complexity.
		\textbf{f}, wind speed distribution.
		\textbf{g}, vehicles distribution.
		\textbf{h}, season distribution.
		\textbf{i}, collection time distribution.}
	\label{fig:Distribution}   
\end{figure*}
\noindent The AVP-IR-Wake dataset is the first large-scale, real-world, physically diverse benchmark for thermal-wake detection. It comprises 161,485 infrared frames collected over three years across multiple lakes, primarily at Tianjin Port (adjacent to the sea), using a cooled infrared camera with a thermal sensitivity of 20~mK and covering multiple underwater vehicle types (Fig.~\ref{fig:Compare} and Fig.~\ref{fig:Distribution}). The dataset captures the extreme conditions under which thermal wakes must be detected. Signal-to-noise ratios range from $-59.18$ to $4.99$~dB, with a median of $-4.09$~dB, and 93.4\% of targets fall below 0~dB, placing the majority of the dataset in the regime of extremely weak signals. Aspect ratios range from 0.159 to 11.492, with a median of 4.497; 9.9\% are below 1 and 90.0\% are above 1, reflecting the elongated geometry of the wake. Target areas span 124.4 to 270,091.4~pixels$^2$, with a median of 48,838.7~pixels$^2$ and 5th, 25th, 75th, and 95th percentiles of 2,902.5, 21,370.2, 84,585.4, and 140,471.0~pixels$^2$, confirming that the wake occupies a large, spatially distributed region rather than a compact object. Target positions show a central bias because the camera orientation was adjusted during acquisition to place most targets near the image center. Background complexity, quantified by variance-weighted information entropy (VIE) \cite{11511399}, has a median of 16,137.4 and a range of 7,467.2 to 22,945.8, with 25th and 75th percentiles of 14,562.1 and 19,080.9, placing most sequences in the high or very high complexity regime (definitions are provided in Supplementary Section~\ref{sec:dataset_metrics}). Environmental diversity is built into the acquisition protocol. Wind speeds span Beaufort 2 to 6, because wind accelerates target dissipation and modulates the cold-skin layer. The data include summer, autumn, and winter in an approximate ratio of 4:4:1, reflecting seasonal changes in cold-skin properties. The ratio of ROV to AUV data is approximately 1:9. Because cold-skin properties also differ between day and night, the data were collected at eight time points throughout the day, with the temporal distribution split between morning (0:00, 2:00, 4:00, 6:00, 8:00) and afternoon (14:00, 16:00, 18:00) at a ratio of 1:2 and uniform sampling within each period. Together, these characteristics establish AVP-IR-Wake as a benchmark that probes the limits of infrared detection: the targets are extremely weak, spatially extended, highly elongated, embedded in complex and variable backgrounds, and observed under a wide range of environmental conditions.

\bigbreak
\noindent
\textbf{Imaging Characteristics of Infrared Thermal Wakes}

\begin{figure*}[!t]
	\centering
	\includegraphics[width=\linewidth]{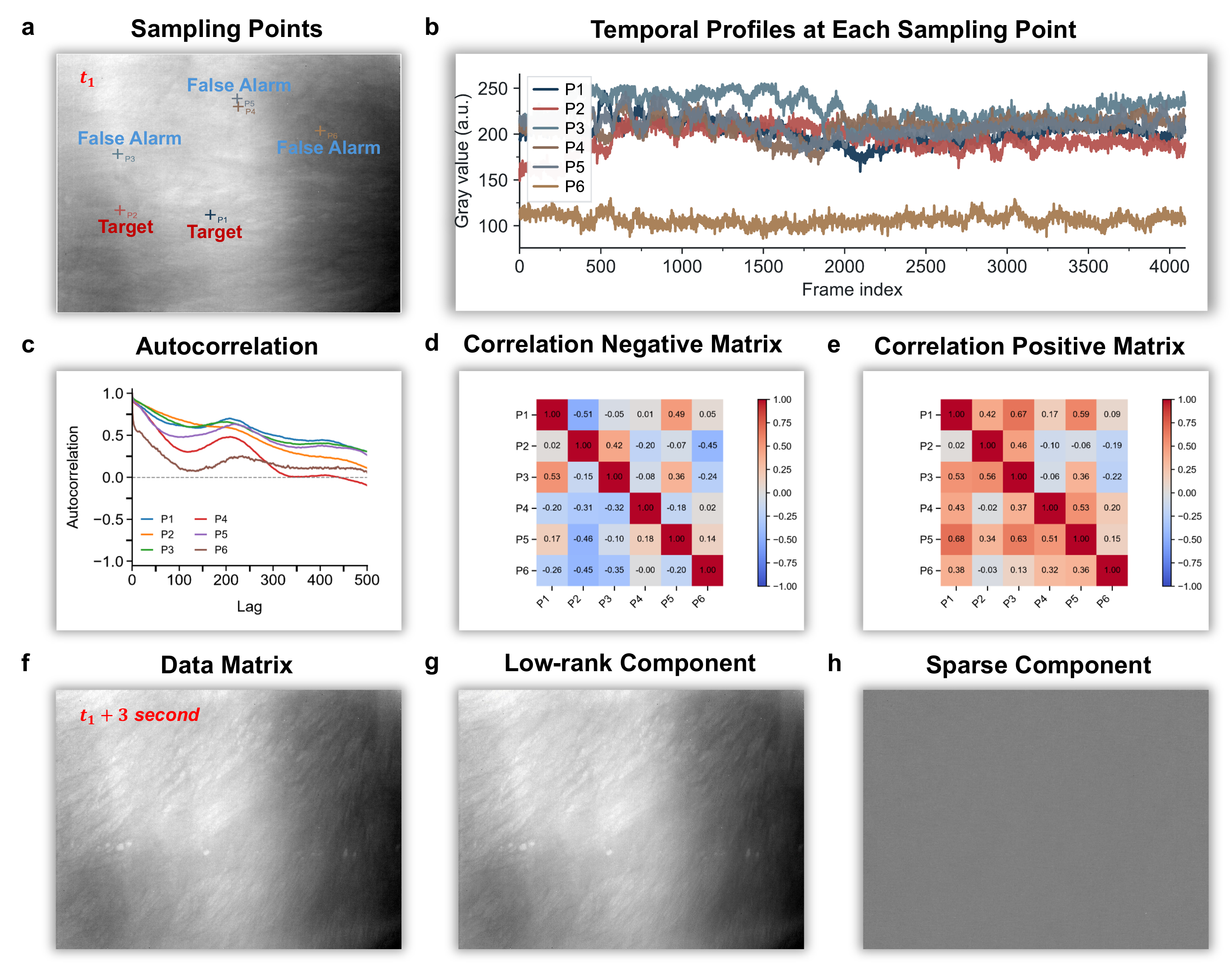}
	\caption{\textbf{Indistinguishability of individual wake vortices and background vortices.} \textbf{a}, Sampling points. P1 and P2 are located on two constituent vortices of the wake; P3 to P6 are located on background false alarms. 
		\textbf{b}, Temporal profiles of pixel intensity at the six sampling points. 
		\textbf{c}, Autocorrelation functions of the six time series. 
		\textbf{d}, Negative time-lagged cross-correlation matrix. 
		\textbf{e}, Positive time-lagged cross-correlation matrix. The correlation between target and background points is comparable to or even higher than that between the two target vortices. 
		\textbf{f},  A representative wake image at \(t_1 + 3\) s. 
		\textbf{g}, Low-rank component obtained by low-rank and sparse decomposition. 
		\textbf{h}, Sparse component. The wake is absorbed into the low-rank component, whereas the sparse component contains no discernible wake structure. 
		These results show that individual wake vortices are not statistically distinguishable from background vortices, with additional results provided in Supplementary Figure \ref{fig:SupTemporalcorrelationanalysis} and \ref{fig:SSupTemporalcorrelationanalysis}, and that the wake does not satisfy the low-rank and sparse assumptions (Further results are provided in Supplementary Figure \ref{fig:SSTemporalcorrelationanalysis}).}
	\label{fig:Temporalcorrelationanalysis}   
\end{figure*}
\textbf{Temporal correlation analysis shows that individual wake vortices are statistically indistinguishable from background vortices in their temporal structure.} As shown in Figure~\ref{fig:Temporalcorrelationanalysis}, we extract pixel intensity time series from six points of interest: two on the constituent vortices of a single thermal wake (P1 and P2) and four on wind-induced background vortices (P3 to P6). Autocorrelation functions for all six points decay slowly and remain well above zero over the full 500-frame window, indicating that both wake and background vortices share long-range temporal persistence. Cross-correlation analysis (Details of the cross-correlation analysis are given in Supplementary Section \ref{sec:Temporalcorrelationanalysis}.) reveals that the target pair reaches its strongest negative correlation of $-0.51$ at a lag of $-483$ frames, but this anti-phase behaviour is not unique: P4 and P5 reach $0.54$ at a lag of 214 frames, and P3 and P5 reach $0.36$ at zero lag. More importantly, the target vortex P1 correlates more strongly with background points P3 and P5 ($0.67$ and $0.59$, respectively) than with its own partner vortex P2 ($0.42$), and the negative correlation between P2 and P6 ($-0.45$) is comparable to that within the target pair. These results show that the temporal signatures of wake vortices are not distinct from those of spontaneously generated background vortices. Both the wake and background structures arise from momentum transport that modulates the surface thermal field \cite{cimoli2026climatic,rai2025atmospheric}, differing only in their triggering mechanism (hydrodynamic displacement versus wind forcing). They therefore share the same spatiotemporal correlation structure. As a further check, low-rank and sparse decomposition \cite{7994669} absorbs the wake into the low-rank component while the sparse component remains nearly empty, indicating that the wake also violates the low-rank and sparse assumptions. Together, these observations establish that the target and background separability assumption fails for thermal wakes: the environment does not merely add noise, but generates structures that are structurally indistinguishable from the target.

\begin{figure*}[!t]
	\centering
	\includegraphics[width=\linewidth]{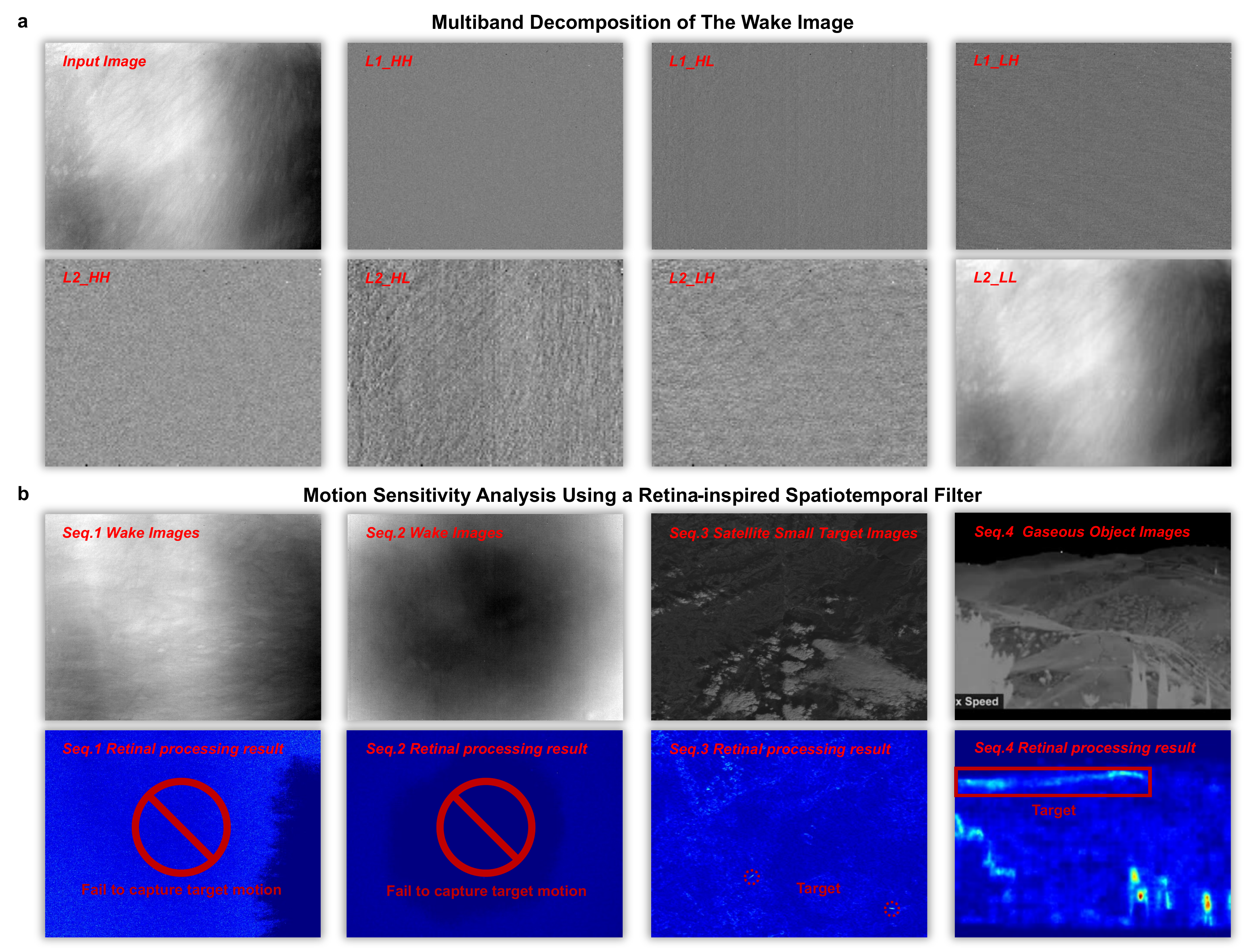}
	\caption{\textbf{Frequency-domain inseparability of thermal wakes and background clutter.} \textbf{a}, Multiband decomposition of a wake image. The wake energy is concentrated in the low-frequency subband (L2\_LL), whereas all high-frequency subbands (L1\_HH, L1\_HL, L1\_LH, L2\_HH, L2\_HL, L2\_LH) contain only background texture and noise. The same low-frequency dominance is observed in background regions, so no spatial-frequency band separates the wake from the clutter (Further results are provided in Supplementary Figure \ref{fig:Sfrequency_inseparability} and Figure \ref{fig:MSE}.). 
	\textbf{b}, Motion sensitivity analysis using a retina-inspired spatiotemporal filter. The filter produces no distinguishable response in wake sequences (Seq.1 and Seq.2), but responds strongly to satellite small targets \cite{10887334} (Seq.3) and gaseous objects \cite{10648775} (Seq.4). This indicates that the wake also lacks a distinctive temporal-frequency signature in the spatiotemporal domain, whereas other targets possess one. 
	Together, these results show that the wake and background clutter share the same low-frequency information in both the spatial and spatiotemporal domains, making frequency-domain separability unattainable.}
	\label{fig:frequency_inseparability}   
\end{figure*}
\textbf{Thermal wakes cannot be separated from background clutter in the frequency domain, in either single-frame or multi-frame settings.} We first examined whether spatial-frequency analysis could isolate the wake from the background. Multiband decomposition of wake images shows that the wake energy is concentrated in the low-frequency subband (L2\_LL), while all high-frequency subbands contain only background texture and noise. Background regions exhibit the same low-frequency dominance, so no spatial-frequency band separates wake from clutter (Figure~\ref{fig:frequency_inseparability}\textbf{a}). This behaviour follows directly from physics: thermal diffusion acts as a low-pass filter on the surface temperature field, suppressing high-frequency components regardless of their origin. We then extended the analysis to the spatiotemporal domain using a retina-inspired filter that responds to frame-to-frame intensity changes. If the wake possessed a distinct temporal-frequency signature, the filter would respond to it; it does not. The filter produces no distinguishable response in wake sequences, whereas it responds strongly to satellite small targets and gaseous objects that possess transient motion (Figure~\ref{fig:frequency_inseparability}\textbf{b}). The wake therefore also lacks a distinctive temporal-frequency signature. Together, these results show that the wake and background clutter occupy the same low-frequency region in both the spatial and spatiotemporal domains, so frequency-domain detectors cannot separate the two. This violates the high-frequency assumption commonly adopted in dim-target and moving-target detection, which presupposes a distinct frequency band for the target.

\begin{figure*}[!t]
	\centering
	\includegraphics[width=\linewidth]{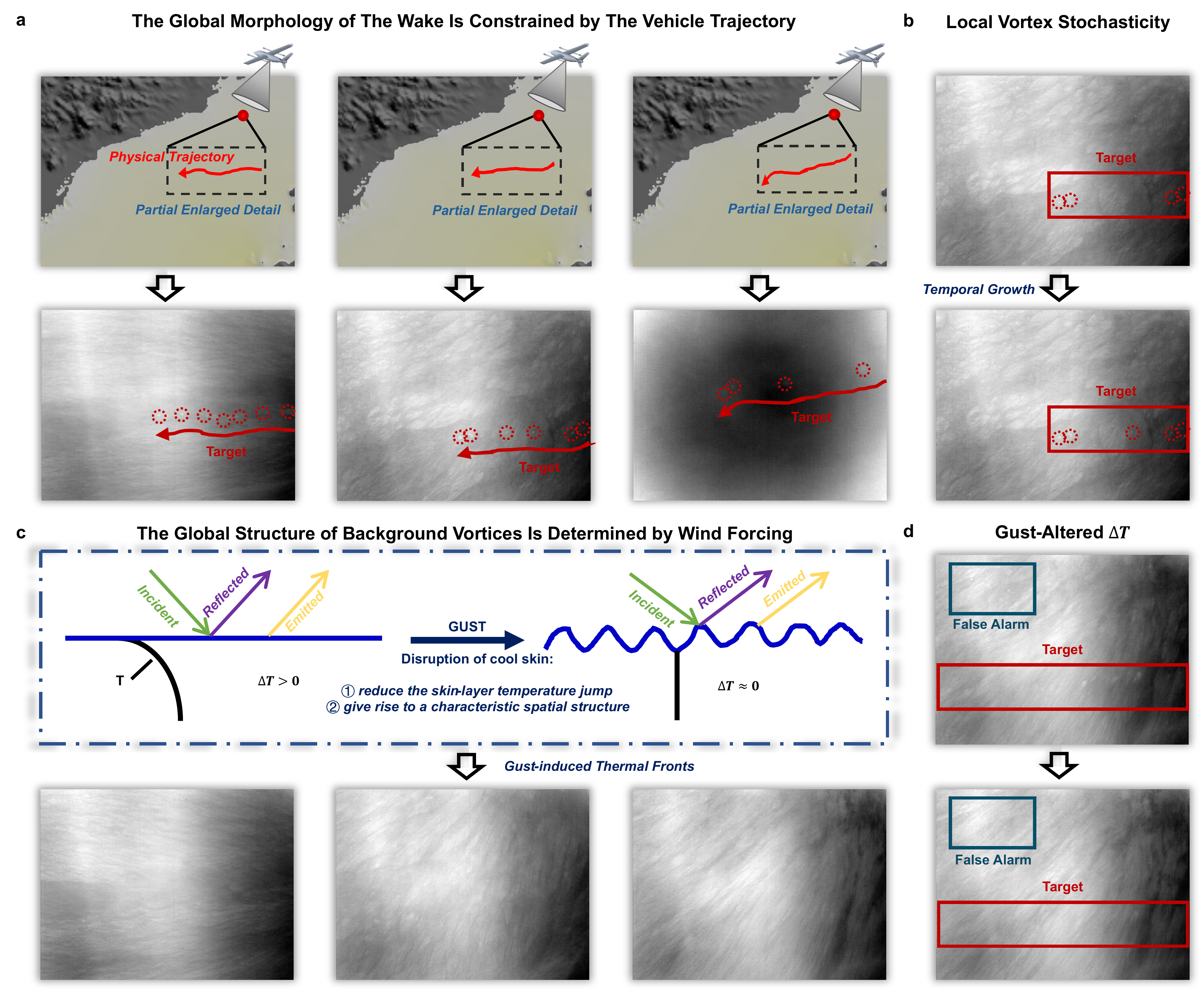}
	\caption{\textbf{Global morphological separability between thermal wakes and background vortices.} \textbf{a}, Wake morphology is trajectory-constrained: individual vortices resemble background vortices, but the overall wake follows the vehicle trajectory (red arrows), forming a distinctive global structure. \textbf{b}, Local vortices appear stochastically in time and space with non-fixed spacing, due to non-uniform vehicle motion. \textbf{c}, Wind generates thermal fronts that organize background vortices into a characteristic global structure. Thus, the wake is organized by the trajectory, whereas background vortices are organized by wind-driven thermal fronts. \textbf{d}, Gusts weaken the cold-skin temperature gradient, reducing both target and background vortex intensities. Because the global trajectory structure is the key to target-background separability, the entire target region was delineated as the bounding box following expert recommendations. This inevitably results in background clutter occupying the dominant spatial proportion within these target boxes. To quantify this, we extracted UMAP \cite{healy2024uniform} embeddings of 130,000 randomly sampled target boxes and corresponding background regions from the same images. The two classes are completely inseparable in the embedding space (Supplementary Fig. \ref{fig:umap}), quantitatively confirming that the visual features within the target boxes are overwhelmingly governed by the surrounding background clutter.}
	\label{fig:Global}   
\end{figure*}
\textbf{The wake remains globally separable from background clutter in its morphology, even though the two are locally indistinguishable.} The preceding analyses show that individual wake vortices cannot be separated from background vortices in either the temporal or the frequency domain. This raises the question of whether any imaging property distinguishes the wake from the background. Figure~\ref{fig:Global}\textbf{a} shows that one does, at the global scale. Although the constituent vortices of the wake are not individually distinguishable from background vortices, the overall extension of the wake follows the physical trajectory of the vehicle (red arrows), yielding a distinctive global morphology that background clutter does not share. A detailed physical analysis is provided in Supplementary Section~\ref{sec:TrajMorph}. This trajectory constraint arises because the surface thermal perturbation accumulates along the vehicle path: the source term is localized around the moving vehicle, and the resulting surface anomaly is the integral of contributions from the vehicle's past positions. The global morphology therefore reflects the trajectory, regardless of local vortex dynamics. Figure~\ref{fig:Global}\textbf{b} further shows that, because underwater vehicles rarely travel at constant speed along a straight line, the constituent vortices appear stochastically in time and space with non-fixed spacing, making the local structure inherently unpredictable. Background vortices exhibit a different global organization. Figure~\ref{fig:Global}\textbf{c} shows that wind forcing generates thermal fronts, and background vortices evolve along these fronts \cite{tsai2023numerical}, forming a characteristic global structure determined by the wind field rather than by any vehicle trajectory. Figure~\ref{fig:Global}\textbf{d} shows that gusts weaken the cold-skin temperature gradient, reducing both target and background vortex intensities. However, because the ocean is thermally stratified, this reduction is depth-dependent. At certain submergence depths, the vehicle's volumetric displacement can still transport warmer subsurface water across the stratified layers to the surface, producing a detectable anomaly even when the ambient cold-skin gradient is weak. In contrast, background vortices generated by wind forcing are confined to the surface layer and are therefore suppressed more strongly. As a result, there exists a depth window in which the target remains observable despite the overall reduction in signal strength. The two classes of structures therefore differ in their global organizing principle: the wake is organized by the vehicle trajectory, whereas background vortices are organized by wind-driven thermal fronts. This global morphological separability is an intrinsic imaging property of thermal wakes. It explains why the wake can be visually distinguished from clutter at large scales, despite being statistically indistinguishable from it at small scales. It also indicates that detection methods operating only on local features cannot exploit this property, because the distinguishing information lies in the global organization of the wake rather than in the local statistics of individual vortices.

\bigbreak
\noindent
\textbf{Quantifying the Breakdown of Conventional Detection Assumptions}

\noindent Directly quantifying how thermal wakes deviate from conventional target assumptions is challenging because infrared wake images are high-dimensional signals shaped by coupled hydrodynamic and environmental processes. Rather than measuring deviations from predefined image statistics, we use existing detection algorithms as quantitative probes of the assumptions embedded in their designs. Algorithmic performance reflects not only model capacity, but also the compatibility between inductive biases and the underlying physical characteristics of the observed targets.
\begin{table*}[htbp]
	\centering
	\vspace{-.03cm}
	\caption{{Quantifying the Breakdown of Conventional Detection Assumptions on AVP-IR-Wake.}}\label{tab-Benchamrk}
	\setlength{\tabcolsep}{1.5mm}
	\begin{tabular}{|clcccccccc|}
		\hline
		&Methods&Venue&$\mathrm{AP}(\%)$&$\mathrm{AP}_{50}(\%)$&$\mathrm{AP}_{75}(\%)$&$\mathrm{AP}_{s}(\%)$&$\mathrm{AP}_{m}(\%)$&$\mathrm{AP}_{l}(\%)$&$\mathrm{AR}(\%)$\\\hline
		\multirow{18}*{\rotatebox{90}{Single-Frame Detector}}
		&DNANet \cite{9864119}  &TIP'23&0.4 &1.5 &0.1 & 0.0 & 0.2 & 0.5 & 14.7\\
		&UIUNet \cite{9989433}  &TIP'23&0.7 &2.7 &0.1 & 0.0 & 0.0 & 1.4 & 3.4\\
		&YOLOH \cite{10471352}  &TIP'24&0.7 &2.8 &0.0 & 0.0 & 0.0 & 0.8 & 4.9\\
		&MSHNet \cite{liu2024infrared}  &CVPR'24&0.0 &0.0 &0.0 & 0.0 & 0.0 & 0.0 & 0.0\\
		&RT-DETR \cite{10657220}  &CVPR'24&4.7 &15.9 &1.4 &0.0 & 0.3 & 5.9 & 30.6\\
		&Hyper-YOLO \cite{10818703} &TPAMI'24&0.5 &2.3 &0.0 & 0.0 & 0.3 & 0.6 & 2.7\\
		&PConv \cite{yang2025pinwheel}  &AAAI'25&0.6 &1.0 &1.0 & 0.0 & 0.0 & 0.6 & 0.0\\
		&LRRNet \cite{11156113} &TGRS'25 &\multicolumn{7}{c|}{ Fail to Achieve Convergence }\\
		&SRSNet \cite{11174084}  &TIP'25&0.2 &1.0 &0.0 & 0.0 & 0.1 & 0.3 & 6.7\\
		&SENet \cite{10844040} &TIP'25&1.2 &3.7 &0.4 & 0.0 & 0.0 & 1.8 & 3.0\\
		&TherNet \cite{11124592} &TPAMI'25&1.4 &4.9 &0.3 & 0.0 & 0.0 & 1.7 & 6.2\\
		&YOLO-MS \cite{10872821} &TPAMI'25&3.2 &13.2 &0.1 & 0.0 & 0.1 & 3.8 & 13.0\\
		&BSR5-DETR \cite{10916506} &TPAMI'25&1.7 &5.8 &0.7 & 0.0 & 1.9 & 2.0 & 27.4\\
		&CSSAM \cite{11297835} &TPAMI'26&1.6 &3.8 &1.7 & 0.1 & 0.0 & 2.8 & 4.5\\
		&$\mathrm{RPCANet}^{++}$ \cite{11537388} &TPAMI'26 &\multicolumn{7}{c|}{ Fail to Achieve Convergence }\\
		&DHiF \cite{11373245} &TCSVT'26&1.2 &4.9 &0.2 & 0.0 & 0.3 & 1.4 & 12.0\\
		&MCFNet \cite{11568953} &TIP'26&1.2 &3.8 &0.7 & 0.0 & 0.1 & 1.8 & 18.0\\
		&NS-FPN \cite{yuan2026seeing} &CVPR'26&0.5 &2.5 &0.0 & 0.0 & 0.0 & 0.6 & 9.4\\
		\hline
		\multirow{20}*{\rotatebox{90}{Multi-Frame Detector}}
		& TransVOD \cite{9960850} & TPAMI'23 &0.3 &1.1 &0.1 & 0.0 & 0.1 & 0.5 & 22.8 \\
		& MICPL \cite{10518061}  & TNNLS'24 &\multicolumn{7}{c|}{ Fail to Achieve Convergence } \\
		& SSTNet \cite{sstnet2024}  & TGRS'24 &0.0 &0.0 &0.0 & 0.0 & 0.0 & 0.0 & 0.0 \\
		& Tridos \cite{10663463}  & TGRS'24 &0.0 &0.0 &0.0 & 0.0 & 0.0 & 0.0 & 0.3 \\
		& ZoomNeXt \cite{10568374}  & TPAMI'24 &1.3 &5.2 &0.1 & 0.0 & 1.5 & 1.5 & 8.5 \\
		& VSF RCNN \cite{10648775} & TPAMI'24 &2.8 &9.3 &1.0 & 0.1 & 0.4 & 3.3 & 12.8 \\
		& HiEUM \cite{10549838}  & TPAMI'24 &\multicolumn{7}{c|}{ Fail to Achieve Convergence } \\
		& DTUM \cite{10321723}  & TNNLS'25 &1.0 &3.5 &0.2 & 0.0 & 0.3 & 1.3 & 12.6 \\
		& EMIP \cite{10989627} & TIP'25 &1.1 &2.1 &1.2 & 0.0 & 0.1 & 1.2 & 5.0 \\
		& OSFormer \cite{11130659}  & TIP'25 &2.5 &9.5 &0.7 & 0.0 & 0.7 & 3.1 & 5.9 \\
		& STDNet \cite{11072100} & TPAMI'25 &1.7 &4.9 &0.1 & 0.0 & 0.0 & 2.0 & 5.5 \\
		& ADSUNet \cite{zhang2026adsunet} & PR'26 &1.7 &4.6 &1.0 & 0.0 & 0.0 & 2.2 & 5.5 \\
		& STGBDNet \cite{11554108} & TGRS'26 &0.8 &1.7 &1.0 & 0.0 & 0.0 & 0.9 & 1.0 \\
		& DQAligner \cite{11363482}  & TGRS'26 &0.0 &0.0 &0.0 & 0.0 & 0.0 & 0.0 & 0.0 \\
		& TDCNet \cite{2026AAAI_TDCNet}  & AAAI'26 &0.0 &0.0 &0.0 & 0.0 & 0.0 & 0.0 & 0.0 \\
		& MISTNet \cite{11511399}  & TIP'26 &0.7 &3.6 &0.1 & 0.1 & 0.5 & 0.9 & 11.6 \\
		& YOLOV++ \cite{shi2026practical} & IJCV'26 &2.6 &9.2 &0.3 & 0.0 & 0.0 & 3.1  & 15.2 \\
		& TenRPCANet \cite{11721414}  & TCSVT'26 &\multicolumn{7}{c|}{ Fail to Achieve Convergence } \\
		& CHAL \cite{Duan_2026_CVPR}  &CVPR'26 &0.3 &1.5 &0.0 & 0.0 & 0.0 & 0.9 & 1.9\\
		& DeepPro \cite{11479915}  &TPAMI'26 &\multicolumn{7}{c|}{ Fail to Achieve Convergence }\\
		\hline
	\end{tabular}
\end{table*}

We evaluated a diverse set of detectors, each configured according to the recommendations in its original publication, and grouped them by the priors they encode. General-purpose single-frame detectors make minimal assumptions about target appearance, while general-purpose multi-frame detectors assume that temporal information aids detection. Infrared dim-target single-frame detectors encode either high-frequency or low-rank/sparse priors, and infrared dim-target multi-frame detectors encode high-frequency, low-rank/sparse, or temporal-profile priors. Because of memory constraints, most multi-frame detectors capture only short-term spatiotemporal relationships spanning about five frames; DeepPro is the exception, processing up to 100 frames by encoding only the temporal profile. Table \ref{tab-Benchamrk} summarizes the results. Across all categories, average precision remains at or near zero, and several methods fail to converge (Supplementary Section \ref{sec:Loss} presents the loss curves of representative detectors. Note that our dataset split was performed following the recommendations of practitioners, and variations in conditions have a substantial impact on both target and background characteristics. Consequently, once a detector fails to capture appropriate representations, its test-set performance generalizes poorly). This systematic failure is not attributable to model capacity or training strategy; rather, it provides a quantitative measure of the mismatch between conventional inductive biases and the physical structure of thermal wakes. Notably, methods that explicitly model low-rank and sparse priors, including LRRNet \cite{11156113}, RPCANet\textsuperscript{++} \cite{11537388}, and TenRPCANet \cite{11721414}, as well as DeepPro \cite{11479915}, which is designed to capture temporal-profile characteristics, all fail to converge. This is consistent with our analyses showing that the low-rank and sparse decomposition assumption is invalidated by the long-range temporal correlations of the wake, and that temporal profiles are locally indistinguishable. CHAL \cite{Duan_2026_CVPR} converges but achieves near-chance performance, because its causal-invariance assumption, which treats target characteristics as independent of the environment, contradicts the environment-dependent nature of thermal wakes, whose physical manifestation evolves together with the surrounding conditions. Furthermore, VSF RCNN \cite{10648775}, a detector designed for gaseous objects, targets highly non-rigid plumes, but its encoded Gaussian dispersion prior fundamentally differs from the physical characteristics of thermal wakes, particularly when the wake consists of discrete vortices, resulting in similarly near-zero performance. Notably, single-frame detectors generally outperform multi-frame detectors; because the spatiotemporal characteristics of thermal wakes are highly complex, and the wake exhibits stochastic local occurrences and long-range temporal correlations, existing multi-frame methods have a fundamental gap between their temporal exploitation strategies (e.g., short-term optical flow or frame differencing) and the actual wake properties, rendering their temporal modeling ineffective. Ultimately, none of the encoded priors, including compactness, high-frequency content, sparsity, temporal profile, and Gaussian dispersion, carries information about target presence in thermal wakes.
\section*{Discussion}
\bigbreak
\noindent
\textbf{Toward Detectors Matched to Thermal Wake Physics}

\noindent The universal failure in Table \ref{tab-Benchamrk} is not a call for larger models or longer training, but for a different class of detector. Current architectures encode priors such as local compactness, high-frequency content, sparsity, and temporal profiles, all incompatible with the physical structure of thermal wakes. A detector suited to this task must instead encode local indistinguishability, stochastic vortex occurrence, low-frequency energy concentration, long-range temporal correlation, and global trajectory-constrained morphology, and must represent the background as wind-organized thermal fronts rather than random noise. This reframes detection from local window classification to global state estimation under physical constraints: the detector tests whether a spatially extended structure consistent with a vehicle trajectory is present, while allowing vortices to appear at random positions and times. Adding global attention to existing architectures is unlikely to suffice, because the required inductive bias is a physical constraint on target organization. Moreover, since background vortices are often associated with wind-driven thermal fronts and wind fields are routinely measured \cite{alet2026operational,achermann2024windseer}, future detectors could incorporate environmental metadata to achieve robust detection beyond imagery alone \cite{11436069}.

\bigbreak
\noindent
\textbf{Implications for Underwater Vehicle Stealth}

\noindent Whereas previous stealth research has focused on controlling cooling-water discharge, a controllable design parameter, our observations demonstrate that thermal wake detectability is governed by the coupled state of the ocean–atmosphere boundary layer. Local indistinguishability implies that reducing the intensity of individual vortices cannot alter their statistical equivalence to background vortices, making vortex-scale suppression ineffective. Global morphological separability implies that the wake follows the vehicle trajectory while background clutter is organized by wind-driven thermal fronts. Because nocturnal cooling stabilizes the cold-skin layer and daytime solar heating weakens it \cite{luo2025modeling}, identical operating conditions are more detectable at night. Because gusts suppress surface-confined background vortices more strongly than subsurface-originating wake signatures, the depth–stealth tradeoff inverts with wind speed: deeper operation is preferable in calm conditions, whereas shallow operation can be safer under strong wind. Stealth is therefore not a fixed capability that can be designed into a platform but a state-dependent property of the ocean–atmosphere system, requiring real-time environmental assessment rather than static design.

\bigbreak
\noindent
\textbf{Underwater Vehicle Observability as An Emergent Property of The Coupled Ocean–Atmosphere System}

\noindent Underwater vehicle observability is governed not only by vehicle-induced hydrodynamic disturbances, but also by the dynamic state of the ocean–atmosphere boundary layer through which these disturbances are transferred into observable signals \cite{huguenin2022drivers,donlon2002toward,minnett2001marine}. Similar to wave breaking \cite{mcallister2024three}, where mechanical energy dissipation modifies the ocean surface skin and regulates exchanges of heat, momentum, and gases between the ocean and atmosphere \cite{deike2022mass}, a submerged vehicle acts as a localized mechanical forcing that perturbs this coupled interface. The resulting infrared wake is therefore not a direct thermal imprint of the vehicle, but an emergent signature shaped by turbulent transport, cold-skin modulation, surface heat exchange, and atmospheric forcing \cite{jessup1997infrared,ford2024enhanced}. Our observations demonstrate that environmental coupling does not simply attenuate wake signatures; it reorganizes the information available for observation, obscuring local thermal structures and spectral characteristics while preserving trajectory-constrained global morphology associated with vehicle motion. Consequently, wake observability should be considered a spatiotemporally varying property of the ocean–atmosphere system rather than an intrinsic attribute of the vehicle itself \cite{waterman2011eddy}. Because the thermal state of the upper ocean varies across geographic regions, seasons, diurnal cycles, and meteorological regimes \cite{sallee2021summertime,su2018ocean,mcgillicuddy2016mechanisms}, the detectability of underwater vehicles is expected to vary accordingly. Future changes in ocean warming, upper-ocean stratification, atmospheric circulation, and extreme weather patterns may further modify these boundary-layer conditions \cite{cheng2022past,li2020increasing,cheng2025ocean}, introducing new challenges for predicting underwater vehicle observability in a changing climate. Addressing these challenges will require interdisciplinary frameworks that couple underwater vehicle hydrodynamics, upper-ocean turbulence, atmospheric forcing, and remote sensing. Oceanographers will need to determine how regional hydrographic states, stratification, and submesoscale processes regulate the persistence of vehicle-induced disturbances; atmospheric scientists will need to quantify how evolving wind regimes, radiative forcing, and air–sea heat exchange modify the thermal transfer pathway; remote sensing communities will need observations capable of resolving fine-scale thermal skin variability across diverse ocean states \cite{donlon2007global}; and computer vision researchers will need physically informed representations that capture evolving global structures rather than relying solely on local target appearance. Such advances will not only improve assessments of underwater vehicle detectability and concealment, but also provide broader insights into how localized mechanical disturbances are transformed into remotely sensed signatures within a changing ocean–atmosphere system.

\bigbreak
\noindent
\textbf{Limitations and Future Work}

\noindent This study focuses on revealing the general structural characteristics and observability patterns of thermal wakes under realistic environmental variability, rather than establishing environment-specific quantitative parameterizations of wake intensity or morphology. Such parameterizations require precise measurements of coupled oceanic, atmospheric, and hydrodynamic variables, which are difficult to obtain in natural environments without sacrificing the environmental diversity essential for identifying general wake characteristics. Moreover, the factors governing wake manifestation vary across regions, seasons, and meteorological conditions, and quantitative relationships derived from limited controlled settings may introduce systematic biases. Therefore, this work emphasizes robust structural representations that remain observable across diverse conditions, providing a foundation for future predictive models of wake observability.
The structural characteristics identified here are derived from a physically diverse dataset collected over three years across multiple lakes, environmental conditions, and vehicle platforms. The consistency of these patterns suggests that wake organization is governed primarily by the underlying cold-skin modulation mechanism and its interaction with the environment, rather than by specific vehicle configurations. Future observations across broader oceanic regimes will be required to establish predictive relationships between ocean–atmosphere states and thermal wake manifestation. Beyond detection, the trajectory-constrained morphology revealed here may enable inverse estimation of vehicle depth, speed, and motion history, transforming thermal wake analysis from binary identification toward quantitative understanding of vehicle–ocean–atmosphere interactions.

\section*{Method}\label{sec:Method}

\subsection*{AVP-IR-Wake Dataset Collection}
\bigbreak
\noindent
\textbf{Operating Envelope Analysis.} A thermal wake is produced when hydrodynamic displacement modulates the cold-skin layer, so not every depth–speed combination gives a detectable wake. To estimate the maximum detectable dive depth, we used the original hydrodynamic model \cite{luo2023thermal,wang2026surface,luo2024reconstruction,luo2024experimental} with the same parameters and settings (we thank Assoc. Prof. Yongcheng Du, who developed this model, for his help), together with the vehicle CAD model, its motion profile, and the camera's 20 mK sensitivity. Under these conditions, the maximum detectable dive depth was $4.5D$, where $D$ is the vehicle diameter.

\bigbreak
\noindent
\textbf{Data Collection.} To ensure sufficient physical and environmental diversity, we conducted extensive field experiments from 2023 to 2025 across multiple inland lakes, primarily at Tianjin Port, as well as inland seas near Zhuhai and Shenzhen, China, using several types of underwater vehicles under diverse operating and environmental conditions. Tianjin Port was selected as the primary site because it more closely approximates the port surveillance scenario for underwater vehicles.
In the experiments, we primarily employed two types of underwater vehicles: remotely operated vehicles (ROVs) and autonomous underwater vehicles (AUVs), which have distinctly different configurations. Among the AUVs, we used the Orange Shark II-T and the Black Shark I-A from Deepinfar. Because these vehicles operate fully submerged, a loss of control would result in sinking. For safety, each vehicle is equipped with an automatic ascent safeguard that triggers upon system failure or upon reaching a preset depth limit. This limit, however, was lower than our calculated \(4.5D\) threshold. We therefore worked with Deepinfar's R\&D team to modify the vehicles and reconfigure their controller parameters accordingly. Vehicle specifications were obtained from the manufacturer's published data (Deepinfar, https://www.deepinfar.com/Cn/About/index.html). 

\subsection*{Data Annotation and Splitting}
\bigbreak
\noindent
\textbf{Data Annotation.}
To annotate the infrared thermal wakes, we developed a physically grounded protocol that integrates multiple complementary priors. The overall wake morphology is self-similar to the vehicle trajectory: straight-line motion produces a straight band, turns produce a curved wake, and hovering collapses the wake into an isolated vortex, while the internal structure remains independent of trajectory shape (Supplementary Section~\ref{sec:TrajMorph}). Wake formation also exhibits a temporal lag, because the volumetric effect requires time to transport subsurface water to the cold-skin layer; this lag increases with depth and decreases with speed, and by recording camera timestamps and vehicle passage times it can be used to reject background false alarms. We further performed CFD simulations to predict the expected image morphology under each depth and speed condition ((Further results are provided in Supplementary Figure \ref{fig:CFD}.)). These predictions guided annotation under most conditions, but when gust-induced thermal fronts aligned with the vehicle motion, the false front closely resembled a real wake, and such cases were resolved through multi-expert cross-validation rather than CFD alone. Because the vehicle trajectory and camera position were logged, the relative position between the camera and the wake provided an auxiliary cue for restricting the search region (Supplementary Section~\ref{sec:CameraWake}). Based on these priors, a panel of experts performed independent annotation and cross-validation, with ambiguous cases reviewed by researchers specializing in thermal-wake fluid dynamics. A gold-standard subset was annotated by experts, and multiple annotators cross-annotated and validated against it, using IoU thresholds of 0.9 for strip-like wakes and 0.8 for vortical targets to reflect their stochastic and discontinuous nature. Annotations meeting these criteria were voted to select the final bounding box, and challenging samples were further reviewed. No videos were excluded on the basis of annotation difficulty or low signal strength.
\bigbreak
\noindent
\textbf{Data Splitting.} Data splitting was performed at the acquisition-day level to prevent data leakage and evaluate generalization to previously unseen operating conditions. Of the 252 collected videos, 52 were excluded (20 for CFD validation and parameter optimization, 32 as redundant or simple), and the remaining 200 videos were split by acquisition-day into training and test sets at an about 3:2 ratio. Notably, the lengths of these video sequences vary. This is because we removed highly similar, redundant frames that occurred under the same operating condition during preprocessing. This redundancy filtering ensures that the dataset captures meaningful temporal dynamics rather than repetitive static states. This design reflects the practical scenario in which a detector must operate across continuously varying environmental states rather than interpolate within a fixed acquisition domain. When learned representations are aligned with the underlying physical mechanisms governing target formation, generalization across such variations is expected to emerge naturally \cite{10648775,li2026generative,du2024conditional}. However, if models rely on environment-specific appearances or inappropriate assumptions about target characteristics, their performance will inevitably degrade when transferred to unseen conditions. Because depth, speed, weather, and wind conditions can vary substantially between acquisition days, this split provides a stringent assessment of whether existing detection paradigms capture physically meaningful and transferable representations or exploit incidental environmental cues (Further results are provided in Supplementary Figure \ref{fig:SDistribution}. An FID score \cite{heusel2017gans} of 252.467 further indicates the substantial distribution discrepancy between the training and test sets, as well as the environment-dependent nature of the targets, with detailed analyses provided in Supplementary Section \ref{sec:FID}.). The proportions of continuous stripe-like targets and discrete vortex-cluster wakes are consistent between the training and test sets, helping ensure that the evaluation is not confounded by differences in the prevalence of these two major wake morphologies.

\subsection*{Frequency-Domain Analysis}
\bigbreak
\noindent
\textbf{Multiband Decomposition.} Spatial-frequency separability was assessed using multiband decomposition based on the two-dimensional discrete wavelet transform. Each infrared frame was converted to grayscale and decomposed using the Haar wavelet with two decomposition levels and symmetric boundary extension. This produced seven subbands: L1\_HH, L1\_HL, L1\_LH, L2\_HH, L2\_HL, L2\_LH, and L2\_LL. The Haar basis was chosen because it provides compact spatial support and is well suited to detecting localized discontinuities without introducing ringing artifacts, which is important for identifying weak and spatially confined features in infrared imagery. Symmetric extension was used to avoid boundary discontinuities that could contaminate the high-frequency subbands.
For each subband, the energy was computed as the sum of squared wavelet coefficients, and the energy fraction was defined as the ratio of that subband's energy to the total energy summed over all subbands. This normalization ensures that the energy fraction is independent of the overall image intensity and directly reflects how the signal is distributed across spatial frequencies. The same decomposition and energy computation were applied to background regions, which were selected as areas of the same size as the target region but located away from the wake.
To verify that the observed low-frequency dominance was not an artifact of the specific wavelet basis or decomposition depth, we repeated the analysis with the Daubechies db2 wavelet and with three decomposition levels. The qualitative result remained unchanged: the wake energy was concentrated in the lowest-frequency subband (L2\_LL or its equivalent), and no spatial-frequency band separated the wake from background clutter.

\bigbreak
\noindent
\textbf{Retina-Inspired Spatiotemporal Filtering.} Retina-inspired filters are advantageous for characterizing spatiotemporal signatures because they emulate the magnocellular pathway’s selective sensitivity to motion and transient intensity changes \cite{gaynes2022classical}. Spatiotemporal separability was assessed using a retina-inspired cellular automaton \cite{liu2026mi} that models the magnocellular pathway of motion processing. The filter operates on consecutive grayscale frames and consists of five layers. The first layer applies photoreceptor adaptation: values above a threshold \(\theta_p = 0.1\) are transformed by \(g_p \tanh(I - \theta_p)\) with \(g_p = 1.5\), while values below are scaled by \(0.1\). The second layer models horizontal cells as \(S_h = \max(S_p - \sigma_h (S_p * K_h), 0)\), where \(K_h\) is a \(3 \times 3\) Gaussian kernel with standard deviation 1.0, \(\sigma_h = 0.3\), and \(*\) denotes convolution. The third layer models bipolar cells with ON and OFF pathways as \(C_t = \max(g_b (S_h - \theta_b), 0) + \max(g_b (-S_h - \theta_b), 0)\), with \(\theta_b = 0.2\) and \(g_b = 2.0\).
The fourth layer models amacrine cells, which accumulate temporal changes. For the first frame, \(R_t = \beta \sqrt{(\nabla_x C_t)^2 + (\nabla_y C_t)^2}\); for subsequent frames, \(R_t = \beta |C_t - S_{\text{prev},b}|\), with \(\beta = 1.2\). The accumulated signal is \(S_a = \alpha S_{\text{prev},a} + (1 - \alpha) R_t\) with \(\alpha = 0.8\). The fifth layer models magnocellular ganglion cells as \(S_m = g_m \max(\tanh(M_s + M_r - \theta_m), 0)\), where \(M_s = I_t * K_m\), \(M_r = \gamma_r S_a\), \(I_t = C_t + \gamma_a S_a\), with \(\gamma_a = 0.5\), \(\gamma_r = 0.7\), \(g_m = 2.5\), \(\theta_m = 0.3\), and \(K_m\) a \(5 \times 5\) Mexican-hat kernel with \(\sigma_1 = 1.0\), \(\sigma_2 = 2.0\), and surround weight 0.5. The final output is \(M_t = \text{Enhance}(\eta_m S_m + (1 - \eta_m) S_a)\), where \(\eta_m = 0.7\) and Enhance applies gamma correction with \(\gamma_p = 0.8\), followed by bilateral filtering and normalization to \([0, 255]\).
The same filter and parameters were applied to thermal wake sequences from AVP-IR-Wake, satellite small-target sequences from IRSatVideo-LEO, and gaseous object sequences from GOD-Video. All sequences were converted to grayscale and processed frame by frame. A response was considered distinguishable if the peak amplitude of \(M_t\) exceeded three times the standard deviation of the background response.

\section*{Code availability}
The code for the temporal profile analysis, single-frame frequency-domain analysis, and retina-inspired spatiotemporal filtering in this work is available at \url{https://github.com/AVP-Group/AVP-IR-Wake-Code}(MIT License). Low-rank and sparse decomposition was performed using the OMoGMF algorithm, available at \url{https://github.com/Yonghongwei/OMoGMF}. The detection algorithms evaluated in this work are available at the following repositories: DNANet at \url{https://github.com/YeRen123455/Infrared-Small-Target-Detection}, UIUNet at \url{https://github.com/danfenghong/IEEE_TIP_UIU-Net}, YOLOH at \url{https://github.com/wsb853529465/YOLOH-main}, MSHNet at \url{https://github.com/BIT-RuiLiu/MSHNet}, RT-DETR at \url{https://github.com/lyuwenyu/RT-DETR}, Hyper-YOLO at \url{https://github.com/iMoonLab/Hyper-YOLO}, PConv at \url{https://github.com/JN-Yang/PConv-SDloss-Data}, LRRNet at
\url{https://github.com/HaLongbao/LRRNet}, SRSNet at \url{https://github.com/fidshu/SRSNet}, SENet at \url{https://github.com/linuxsino/SENet}, TherNet at \url{https://github.com/mzzcdf/TherNetV1}, YOLO-MS at \url{https://github.com/FishAndWasabi/YOLO-MS}, CSSAM at \url{https://github.com/ettof/CSOD}, BSR5-DETR at \url{https://github.com/H1kari06},  RPCANet++ at \url{https://github.com/fengyiwu98/RPCANet}, DHiF at \url{https://github.com/TinaLRJ/DHiF}, MCFNet at \url{https://github.com/Fihven/MCFNet}, NS-FPN at \url{https://github.com/mengduann/NS-FPN}, TransVOD at \url{https://github.com/SJTU-LuHe/TransVOD}, MICPL at \url{https://github.com/UESTC-nnLab/MICPL}, SSTNet at \url{https://github.com/UESTC-nnLab/SSTNet}, Tridos at \url{https://github.com/UESTC-nnLab/Tridos}, ZoomNeXt at \url{https://github.com/lartpang/ZoomNeXt}, VSF RCNN at \url{https://github.com/CalayZhou}, DTUM at \url{https://github.com/TinaLRJ}, EMIP at \url{https://github.com/zhangxin06/EMIP}, OSFormer at \url{https://github.com/q2479036243/OSFormer}, STDNet at \url{https://github.com/hellozhuo/stdnet}, ADSUNet at \url{https://github.com/zhanglw882/ADSUNet}, STGBDNet at \url{https://github.com/greekinRoma}, DQAligner at \url{https://github.com/dengfa02/DQAligner_MIRSTD}, TDCNet at \url{https://github.com/IVPLabs/TDCNet}, MISTNet at \url{https://github.com/ShuCvlab/MIST}, YOLOV++ at \url{https://github.com/YuHengsss/YOLOV}, CHAL at \url{https://github.com/UESTC-nnLab/CHAL}, and DeepPro at \url{https://github.com/TinaLRJ/DeepPro}.

\section*{Data availability}
The full AVP-IR-Wake dataset cannot be made publicly available; however, access can be requested from the first author. This complies with Nature’s data availability requirements (https://www.nature.com/documents/nr-data-availability-statements-data-citations.pdf). Its construction involved parameters provided by multiple Chinese research institutions, and underwater vehicle detection is a sensitive research topic. The vehicle models used in the dataset were primarily provided by Deepinfar, and some of their characteristics are related to the company's products. The confidentiality office of Sun Yat-sen University has reviewed the dataset and determined that it does not contain information subject to any confidentiality classification, but that it is restricted to internal use. Access to the dataset can be granted upon application and approval by the participating institutions. Researchers seeking access may contact the first author at zhanggy57@mail2.sysu.edu.cn. The first author will coordinate the internal review and authorization process with the participating institutions on behalf of the applicant. Requests should be submitted in writing and should include: (i) the intended research purpose; (ii) the applicant's institutional affiliation; and (iii) a commitment that the data will not be used for commercial purposes. The dataset has also been deposited in the data center of the China Ship Research and Design Center; authorized personnel of that center may access it directly. The outcome of each access request will be communicated within 60 days at the latest. The CFD temperature fields and the physical model of wake formation are not included in this data availability statement. Their generation involved parameters provided by multiple institutions, and the intellectual property of the wake formation model belongs to the Institute of Vibration and Noise, Naval University of Engineering. Researchers interested in these components should contact that institute directly for access. The data underlying the figures in the main text have been submitted as part of the manuscript. Additionally, we provide a supplementary video (\textit{``Supplementary Video 1.avi''}) to offer editors, reviewers, and readers a direct and intuitive visual understanding of the thermal wake imagery. This video presents a short, representative clip of the real-world infrared recording rather than the complete evolution process, serving to complement the static frames shown in the main text.

Two external datasets were used for the comparative analysis in Fig.~\ref{fig:frequency_inseparability}\textbf{b} to illustrate the contrast between thermal wakes and other classes of infrared targets. The infrared satellite small-target images are from the IRSatVideo dataset (IRSatVideo-LEO), a semi-simulated benchmark for multi-frame infrared small-target detection in satellite videos, comprising 200 image sequences and 91,021 frames developed from Landsat 7–8 imagery. The dataset is available upon request via the form at \url{https://github.com/XinyiYing/IRSatVideo}. The gaseous object images are from the GOD-Video dataset, a benchmark for gaseous object detection comprising 600 videos and 141,017 frames covering multiple gas types. The GOD-Video dataset is licensed under the Creative Commons Attribution-NonCommercial 4.0 International (CC BY-NC 4.0) license, and access is granted upon request via the dataset website at \url{https://calayzhou.github.io/2022/05/04/Dataset}.

\section*{Acknowledgement}
This work received no external funding. We thank the Institute of Vibration and Noise, Naval University of Engineering, and the National Key Laboratory on Ship Vibration and Noise for their assistance, and in particular Associate Professor Yongcheng Du, Dr. Feiyang Luo, and Dr. Chengzhe Gao for their help and discussions on the fluid physics of wake formation. We also thank Senior Engineers Zongliang Bai, Zhiwen Leng and Zhihao Tang from the China Ship Research and Design Center for their discussions and suggestions on the AVP-IR-Wake dataset splitting based on practical usage scenarios. We are also grateful to Researcher Jun Zhang from the China Ship Scientific Research Center for his valuable discussions on propeller-induced mixing of underwater vehicles. We also thank Associate Professor Cheng'an Wang from the School of Energy Science and Engineering, Harbin Institute of Technology, for his valuable discussions and assistance on the conversion from temperature fields to radiation fields, and for providing the simulated images generated from our CFD-computed temperature fields. We further thank the Shanghai Institute of Technical Physics, Chinese Academy of Sciences, for its assistance with the design of experimental conditions, and in particular Professor Biao Chen for his discussions. We are also grateful to Senior Engineer Liang Peng and Chao Zuo from the Wuhan Second Ship Design and Research Institute, Researcher Zhiben Shen from the Hanjiang National Laboratory, and Researcher Haigang Ren from the Systems Engineering Research Institute of CSSC for their assistance and valuable suggestions. We are also grateful to Professor Yang Shang from the College of Aerospace Science and Engineering, National University of Defense Technology, for his valuable suggestions on data collection. We also thank Professor Yueqiang Zhang from the College of Physics and Optoelectronic Engineering, Shenzhen University, and Professor Yan Zhang from the School of Electronics and Information Technology, Sun Yat-sen University, for their assistance. We also thank Associate Professor Yunxiao Shan from the School of Artificial Intelligence, Sun Yat-sen University, for his assistance with data collection. Furthermore, we acknowledge Guoyi Zhang and Xiaohu Zhang for their dedicated leadership and full involvement in conducting this thermal wake project from 2023 to 2025, and we appreciate the contributions of the many co-authors who provided assistance at various stages of the project.  We also thank Deepinfar for providing its R\&D and engineering experiment teams to assist with the lake trials and the collection of the AVP-IR-Wake dataset.

\section*{Author contributions}
Guoyi Zhang: Conceptualization, Methodology, Formal analysis, Investigation, Data curation, Validation, Visualization, Writing, original draft, Writing, review \& editing. Yanjin Du: Data curation, Writing, review \& editing. Guangsheng Xu: Data curation, Writing, review \& editing. Zhengyao Zhao: Data curation, Writing, review \& editing. Siyang Chen: Data curation, Writing, review \& editing. Jingjing Gao: Data curation, Writing, review \& editing. Yuezheng Zhou: Data curation, Writing, review \& editing. Yiyang Wu: Data curation, Writing, review \& editing. Bin Wu: Data curation, Writing, review \& editing. Tongsu Zhang: Data curation, Writing, review \& editing. Lejia Ye: Data curation, Writing, review \& editing. Donghe Wang: Data curation, Writing, review \& editing. Jie Wang: Data curation, Writing, review \& editing. Kaiye Hu: Data curation, Writing (fluid physics), review \& editing. Zhipeng Ren: Supervision, Writing, review \& editing. Han Wang: Supervision, Writing, review \& editing. Boo Cheong Khoo: Supervision, Writing, review \& editing. Xiangpeng Xu: Supervision, Writing, review \& editing. Xiaohu Zhang: Supervision, Writing, review \& editing, Funding acquisition.

\section*{Competing interests}
The authors declare that they have no known competing financial interests or personal relationships that could have appeared to influence the work reported in this paper.

\clearpage

\newpage
\vspace*{\baselineskip}
\begin{center}
	{\Large\bfseries Supplementary Information}
\end{center}
\vspace*{\baselineskip}
\begin{center}
	{\Large\bfseries Revealing Imaging Characteristics of Thermal Wakes from Underwater Vehicles}\\[4pt]
	{\Large\bfseries  Without Cooling-Water Discharge}
\end{center}

\setcounter{section}{0}
\renewcommand{\thesection}{S\arabic{section}}
\setcounter{figure}{0}
\renewcommand{\thefigure}{S\arabic{figure}}
\setcounter{table}{0}
\renewcommand{\thetable}{S\arabic{table}}

\normalsize

\section{Comparison with Simulated Imaging}
\phantomsection
\label{sec:simulation_comparison}
Synthetic infrared imaging has been widely used to study thermal wakes, typically relying on two distinct strategies: direct computational simulation and target embedding. Neither strategy can substitute for real-world observations when evaluating detection algorithms.

\begin{figure*}
	\centering
	\includegraphics[width=\linewidth]{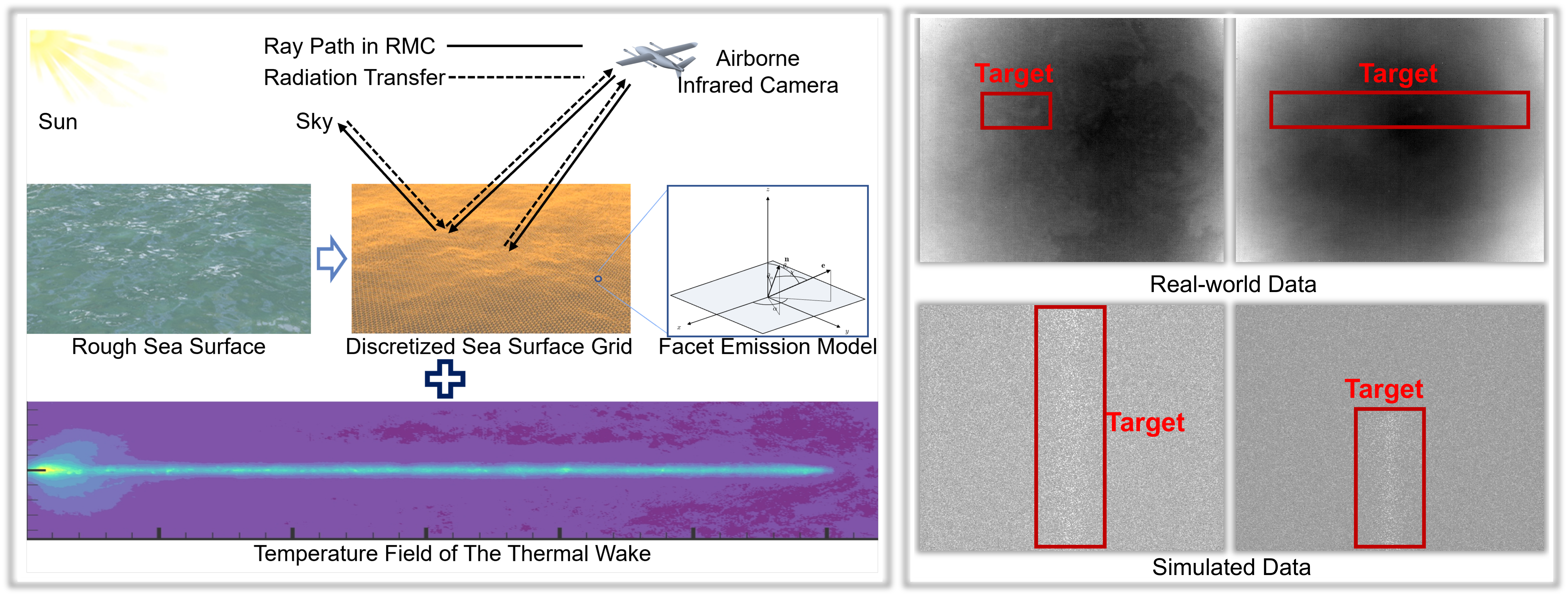}
	\caption{\textbf{Comparison with Simulated Imaging.} Simulated images can be computed from temperature fields and a rough sea surface emission model by reverse Monte Carlo (RMC); however, these data fail to capture the complex environmental information necessary to establish a realistic background. We thank Associate Professor Cheng'an Wang for providing the simulated data and computational results.}
	\label{fig:Background}   
\end{figure*}

The first strategy generates images by converting Computational Fluid Dynamics (CFD) temperature fields directly into radiance fields. While this approach accurately reproduces the idealized physical morphology of the wake, it fails to capture the complex environmental coupling, as shown in Figure \ref{fig:Background}. Real ocean surfaces are continuously modulated by wind, waves, and solar radiation, generating thermal fronts and background clutter. Since the fundamental challenge of target detection lies in separating the target from this complex background, computationally simulated images lack the realistic environmental interference necessary to evaluate a detector's true discriminative capability. 

The second strategy attempts to preserve realistic backgrounds by embedding synthetic wake signatures into real infrared scenes. However, this approach violates physical consistency. In reality, the background state (e.g., wind speed, wave roughness, and cold-skin temperature gradient) is physically coupled with the wake generation process. A wake observed under strong wind conditions possesses different morphological and thermal characteristics than one generated under calm seas. Directly overlaying a wake patch onto an arbitrary background ignores this coupling. Furthermore, this embedding approach is constrained by the lack of high-sensitivity, long-term infrared observations of the ocean surface required to serve as high-fidelity background templates.

Consequently, neither simulation strategy provides a valid proxy for algorithm benchmarking. The idealized targets of computational simulations and the physically inconsistent backgrounds of embedded datasets fail to reflect the local indistinguishability between wakes and natural clutter observed in our real-world dataset. This underscores why AVP-IR-Wake, acquired entirely through high-sensitivity field observations, is indispensable for characterizing the true imaging properties of thermal wakes.

\section{Metrics}
\phantomsection
\label{sec:dataset_metrics}

\subsection{Dataset Metrics}
The difficulty and physical diversity of AVP-IR-Wake are mainly characterized by two metrics: signal-to-noise ratio (SNR) and variance-weighted information entropy (VIE). SNR is defined as
\begin{equation}
	\mathrm{SNR} = 10\lg\left(\frac{|\mu_t - \mu_b|}{\sigma_b}\right),
\end{equation}
where $\mu_t$ denotes the mean intensity of the target region, and $\mu_b$ and $\sigma_b$ denote the mean and standard deviation of the local background region, respectively.

For the $t$-th frame $\mathbf{I}_t$ in a sequence, its VIE is defined as
\begin{equation}
	\mathrm{VIE} = -\sum_{\beta=0}^{255} \left(\beta - \bar{\beta}_t\right)^2 p(\mathbf{I}_t = \beta) \log p(\mathbf{I}_t = \beta),
\end{equation}
where $\beta$ denotes the grayscale variable in the histogram, $p(\mathbf{I}_t = \beta)$ is the probability of grayscale value $\beta$ occurring in frame $\mathbf{I}_t$, and $\bar{\beta}_t$ is the mean grayscale value of $\mathbf{I}_t$. The VIE metric thus quantifies the fluctuation of the grayscale distribution. Based on the sequence’s average VIE, the background complexity is divided into four levels: low (0--1,000), medium (1,000--5,000), high (5,000--10,000), and very high (>10,000).

\subsection{Evaluation Metrics}
We report standard COCO-style detection metrics: AP, AP\(_{50}\), AP\(_{75}\), AP\(_{s}\), AP\(_{m}\), AP\(_{l}\), and AR. AP is the average precision, computed as the area under the precision--recall curve, averaged over IoU thresholds from 0.50 to 0.95 in steps of 0.05. AP\(_{50}\) and AP\(_{75}\) are the AP values at IoU thresholds of 0.50 and 0.75, respectively. AP\(_{s}\), AP\(_{m}\), and AP\(_{l}\) denote AP for small, medium, and large objects, defined by object area following the COCO convention: small (area \(< 32^2\) pixels), medium (\(32^2 \le\) area \(\le 96^2\) pixels), and large (area \(> 96^2\) pixels). AR is the average recall, computed as the maximum recall at a fixed number of detections per image, averaged over IoU thresholds. 

\section{Temporal correlation analysis}
\phantomsection
\label{sec:Temporalcorrelationanalysis}
For each point of interest, we extract the pixel intensity time series \(x_t\) from the infrared image sequence, where \(t = 1, \ldots, N\). The temporal mean is denoted by \(\bar{x} = N^{-1}\sum_{t=1}^{N} x_t\). The autocorrelation function \(\rho(l)\) at lag \(l\) (\(l = 0, 1, \ldots, L\)) is defined as
\begin{equation}
	\rho(l) = \frac{\sum_{t=1}^{N-l} (x_t - \bar{x})(x_{t+l} - \bar{x})}{\sum_{t=1}^{N} (x_t - \bar{x})^2},
	\label{eq:autocorr}
\end{equation}
with \(\rho(0) = 1\). For two time series \(x_t\) and \(y_t\), the cross-correlation function \(\rho_{xy}(l)\) at lag \(l\) (\(l = -L, \ldots, L\)) is defined as
\begin{equation}
	\rho_{xy}(l) = 
	\begin{cases}
		\displaystyle \frac{\sum_{t=1}^{N-l} (x_t - \bar{x})(y_{t+l} - \bar{y})}
		{\sqrt{\sum_{t=1}^{N}(x_t - \bar{x})^2 \sum_{t=1}^{N}(y_t - \bar{y})^2}}, & l > 0, \\[10pt]
		\displaystyle \frac{\sum_{t=1}^{N} (x_t - \bar{x})(y_t - \bar{y})}
		{\sqrt{\sum_{t=1}^{N}(x_t - \bar{x})^2 \sum_{t=1}^{N}(y_t - \bar{y})^2}}, & l = 0, \\[10pt]
		\displaystyle \frac{\sum_{t=1}^{N+l} (x_{t-l} - \bar{x})(y_t - \bar{y})}
		{\sqrt{\sum_{t=1}^{N}(x_t - \bar{x})^2 \sum_{t=1}^{N}(y_t - \bar{y})^2}}, & l < 0.
	\end{cases}
	\label{eq:crosscorr}
\end{equation}

To summarize pairwise correlations, we construct a positive-correlation matrix \(M^{+}\) and a negative-correlation matrix \(M^{-}\), with elements
\begin{equation}
	M^{+}_{ij} = \max_{l>0} \rho_{x_i x_j}(l), \qquad
	M^{-}_{ij} = \min_{l<0} \rho_{x_i x_j}(l),
	\label{eq:matrices}
\end{equation}
where \(x_i\) and \(x_j\) denote the time series at points \(i\) and \(j\), respectively. Here \(L\) is the maximum lag considered, set to \(L = 500\) in our analysis. All time series are mean-subtracted prior to correlation analysis.

\begin{figure*}[!t]
	\centering
	\includegraphics[width=\linewidth]{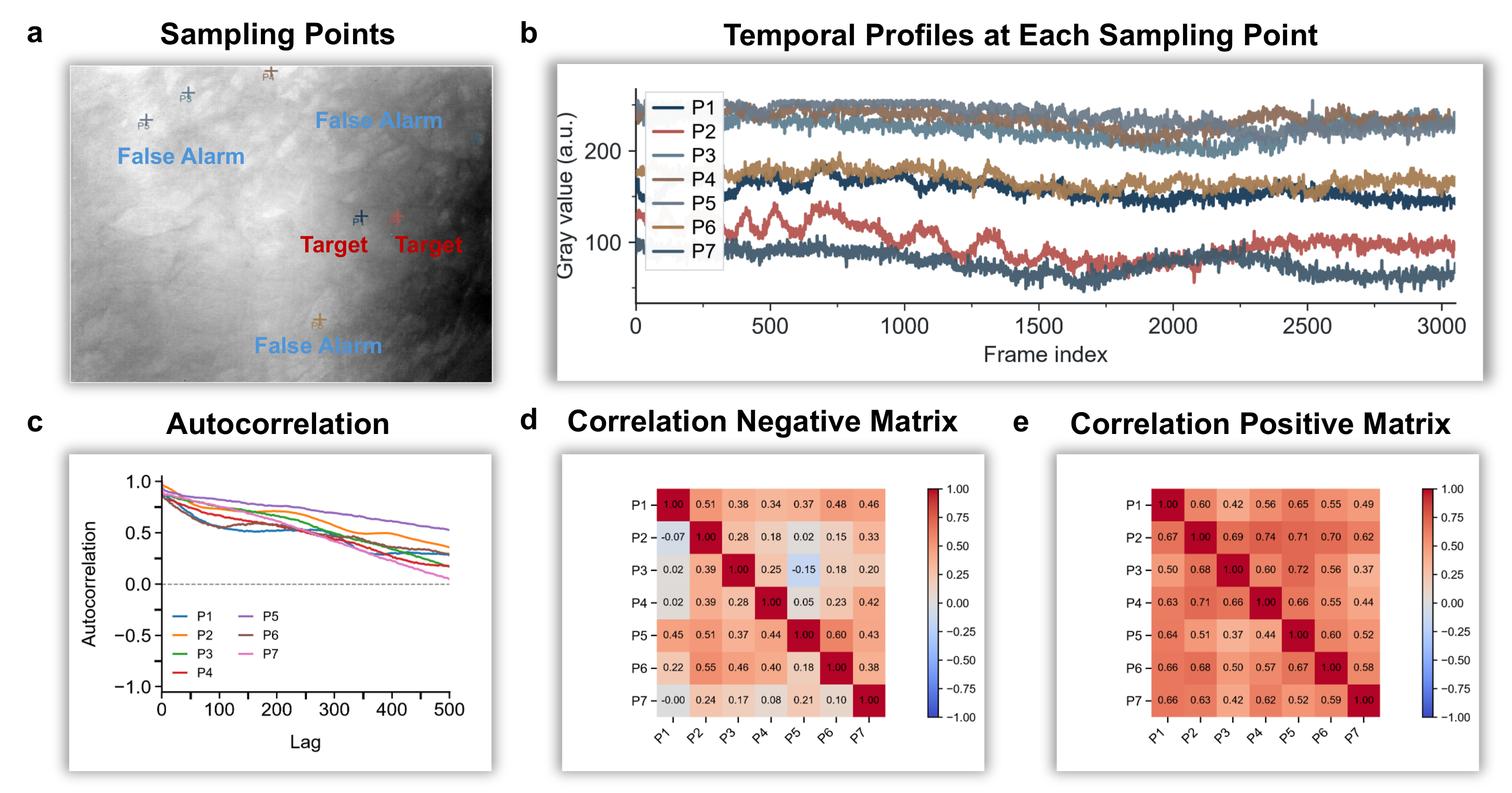}
	\caption{\textbf{Indistinguishability of individual wake vortices and background vortices.} \textbf{a}, Sampling points. P1 and P2 are located on two constituent vortices of the wake; P3 to P7 are located on background false alarms. 
		\textbf{b}, Temporal profiles of pixel intensity at the seven sampling points. 
		\textbf{c}, Autocorrelation functions of the seven time series. 
		\textbf{d}, Negative time-lagged cross-correlation matrix. 
		\textbf{e}, Positive time-lagged cross-correlation matrix. The correlation between target and background points is comparable to or even higher than that between the two target vortices.}
	\label{fig:SupTemporalcorrelationanalysis}   
\end{figure*}

\begin{figure*}[!t]
	\centering
	\includegraphics[width=\linewidth]{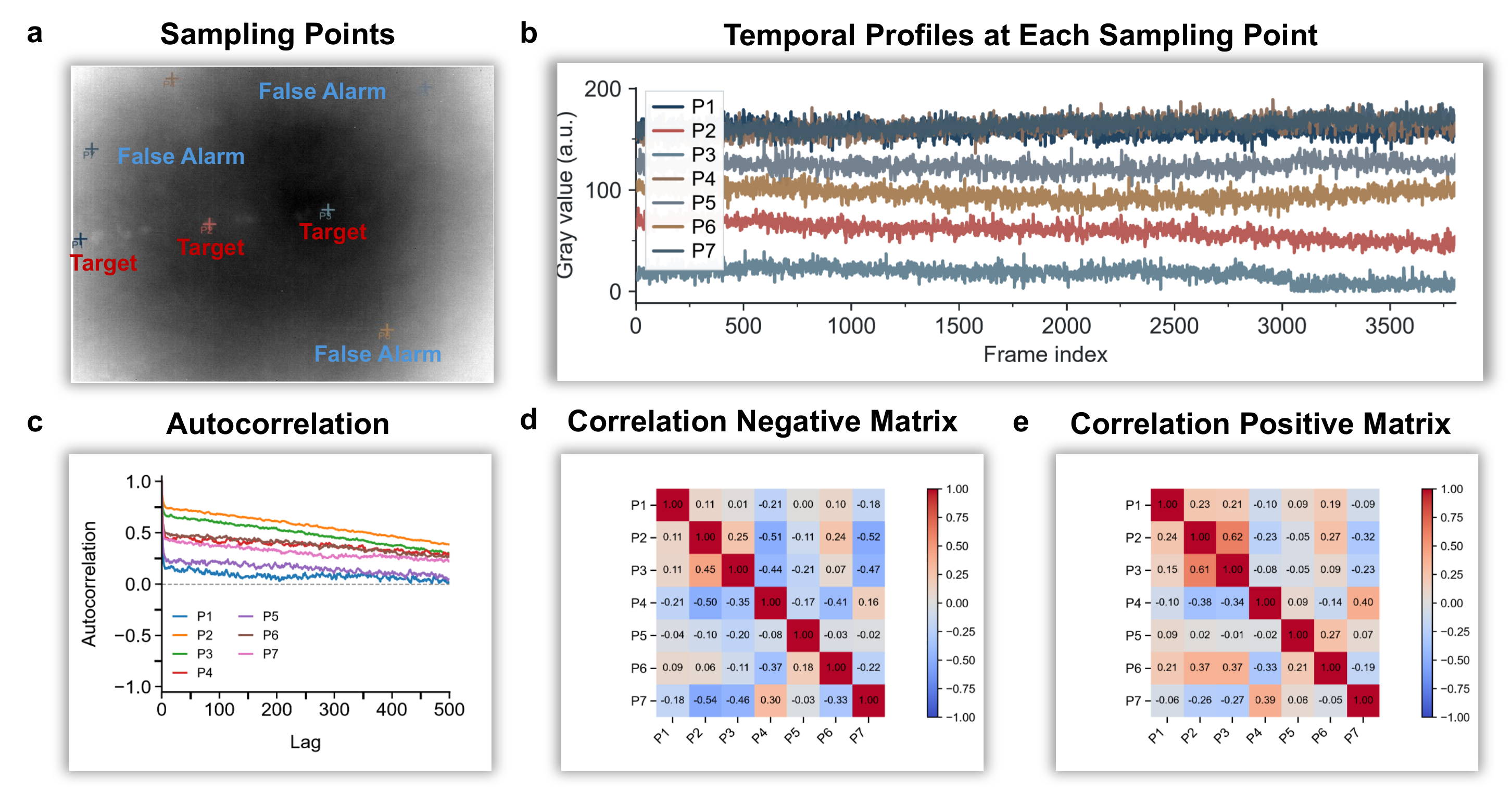}
	\caption{\textbf{Indistinguishability of individual wake vortices and background vortices.} \textbf{a}, Sampling points. P1 to P3 are located on two constituent vortices of the wake; P4 to P7 are located on background false alarms. 
		\textbf{b}, Temporal profiles of pixel intensity at the seven sampling points. 
		\textbf{c}, Autocorrelation functions of the seven time series. 
		\textbf{d}, Negative time-lagged cross-correlation matrix. 
		\textbf{e}, Positive time-lagged cross-correlation matrix. The correlation between target and background points is comparable to or even higher than that between the two target vortices.}
	\label{fig:SSupTemporalcorrelationanalysis}   
\end{figure*}

\begin{figure*}[!t]
	\centering
	\includegraphics[width=\linewidth]{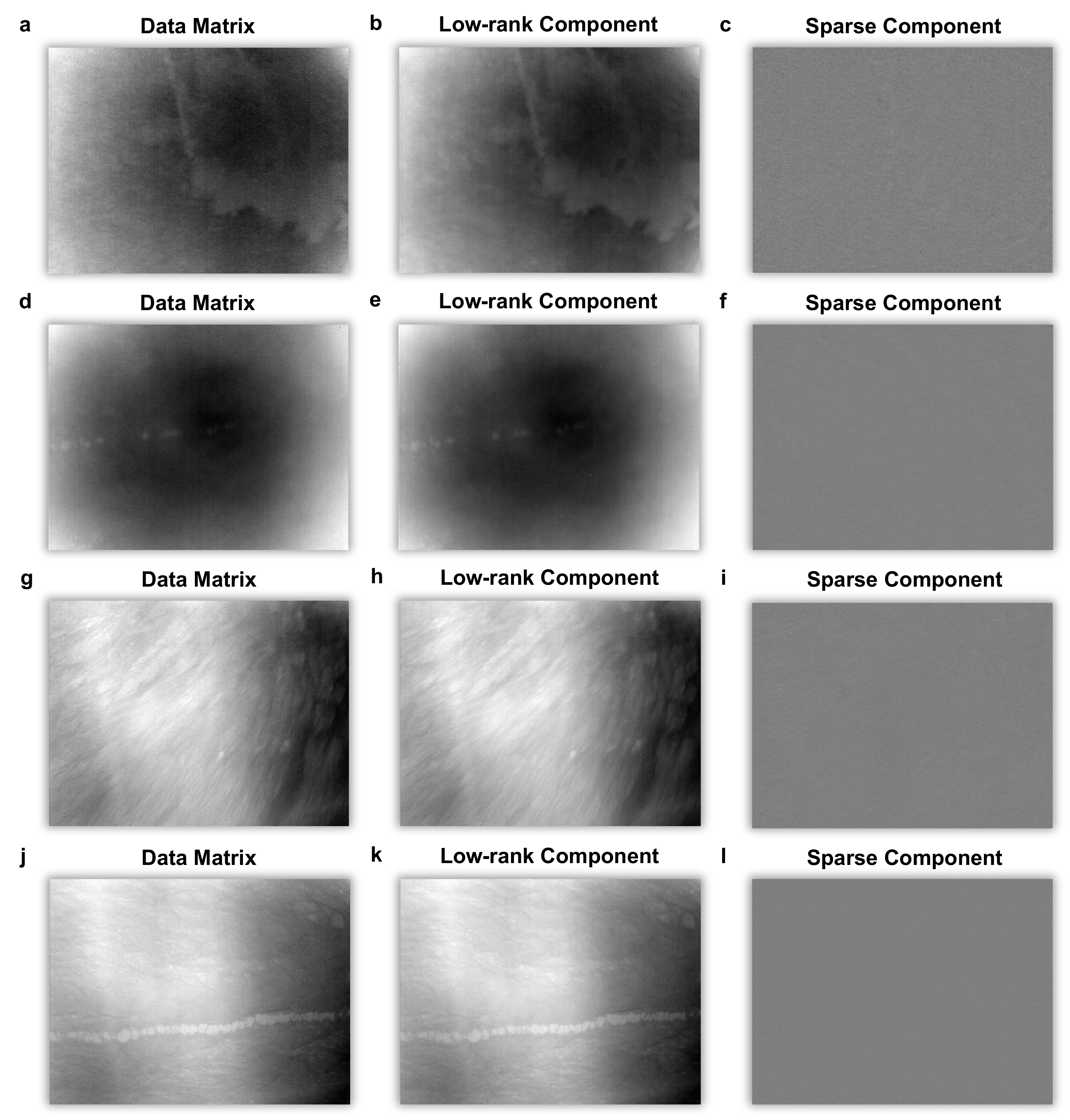}
	\caption{\textbf{The wake does not satisfy the low-rank and sparse assumptions.}}
	\label{fig:SSTemporalcorrelationanalysis}   
\end{figure*}

\begin{figure*}[!t]
	\centering
	\includegraphics[width=\linewidth]{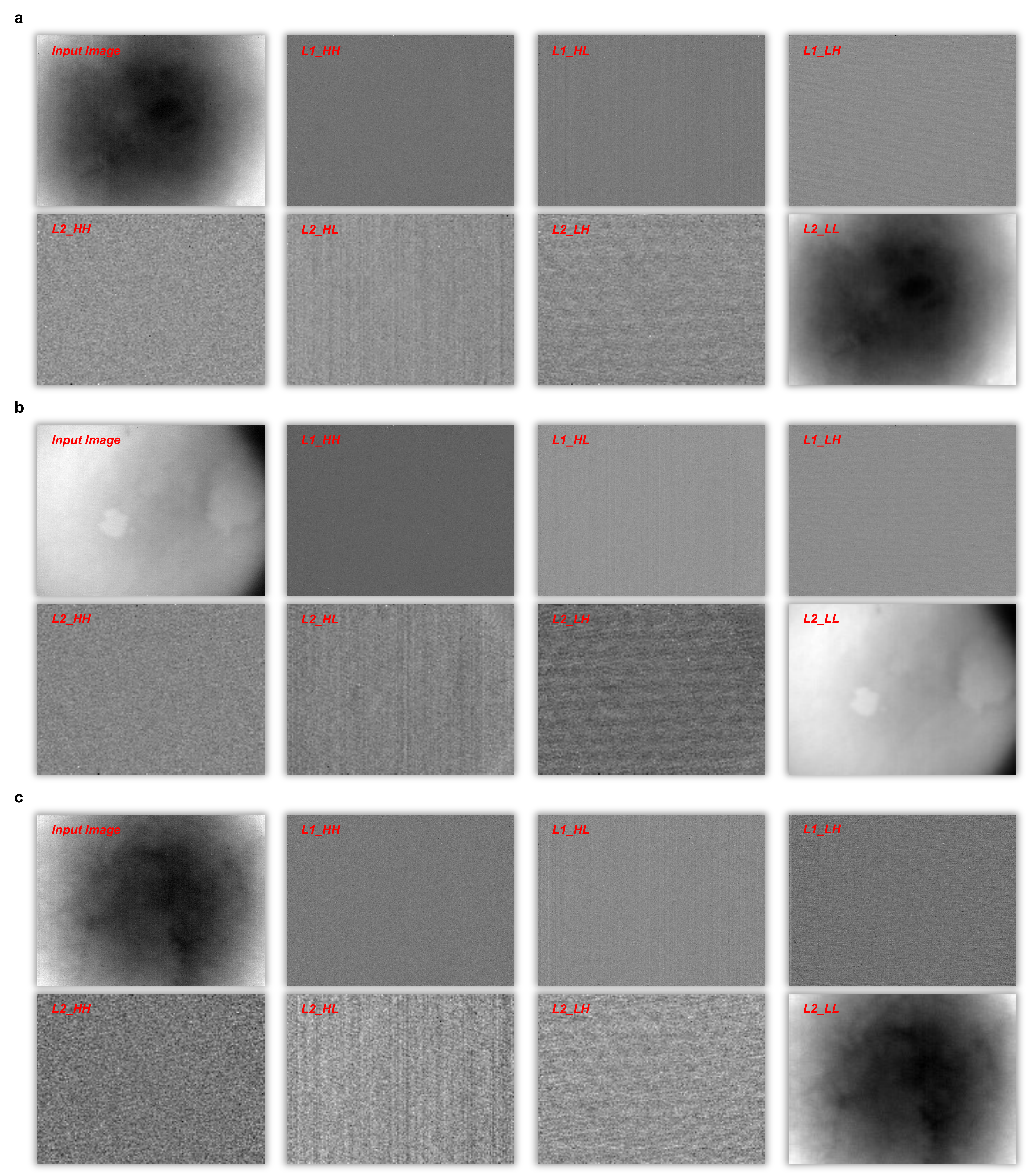}
	\caption{\textbf{ Multiband decomposition of a wake image.} The wake energy is concentrated in the low-frequency subband (L2\_LL), whereas all high-frequency subbands (L1\_HH, L1\_HL, L1\_LH, L2\_HH, L2\_HL, L2\_LH) contain only background texture and noise. The same low-frequency dominance is observed in background regions, so no spatial-frequency band separates the wake from the clutter.}
	\label{fig:Sfrequency_inseparability}   
\end{figure*}

\begin{figure*}[!t]
	\centering
	\includegraphics[width=\linewidth]{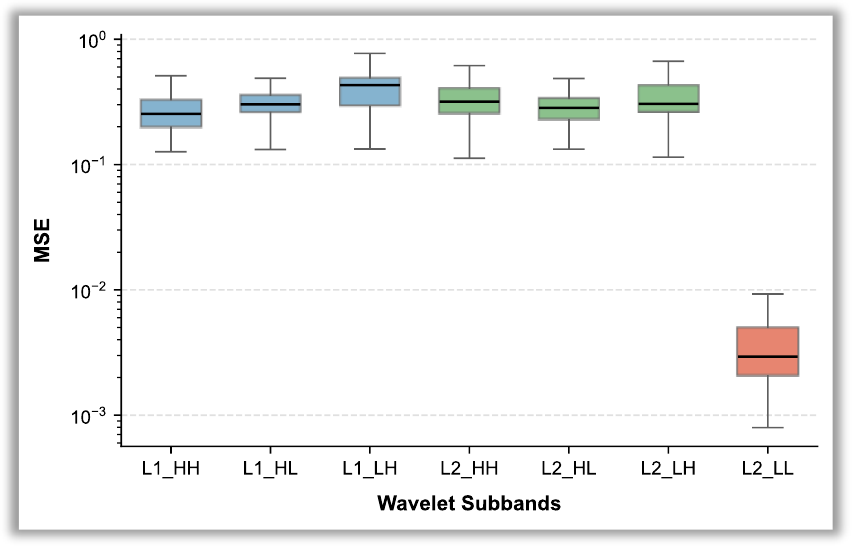}
	\caption{\textbf{Relative frequency-domain energy distribution across the entire dataset.} Boxplots showing the Mean Squared Error (MSE) between the original images and their respective wavelet subbands across the entire 161,485-frame dataset. To facilitate comparison, the MSE values are normalized by the global maximum MSE across all frames and subbands. The y-axis is presented on a logarithmic scale. Even after this normalization, the lowest-frequency subband (L2\_LL) exhibits an MSE nearly two orders of magnitude lower than the high-frequency detail subbands. This confirms that the image energy is overwhelmingly concentrated in the lowest frequency band, while high-frequency bands contain only noise. Boxes represent the interquartile range  (IQR, 25th–75th percentiles), center lines indicate the median, and whiskers extend to 1.5× IQR.}
	\label{fig:MSE}   
\end{figure*}

\begin{figure*}[!t]
	\centering
	\includegraphics[width=\linewidth]{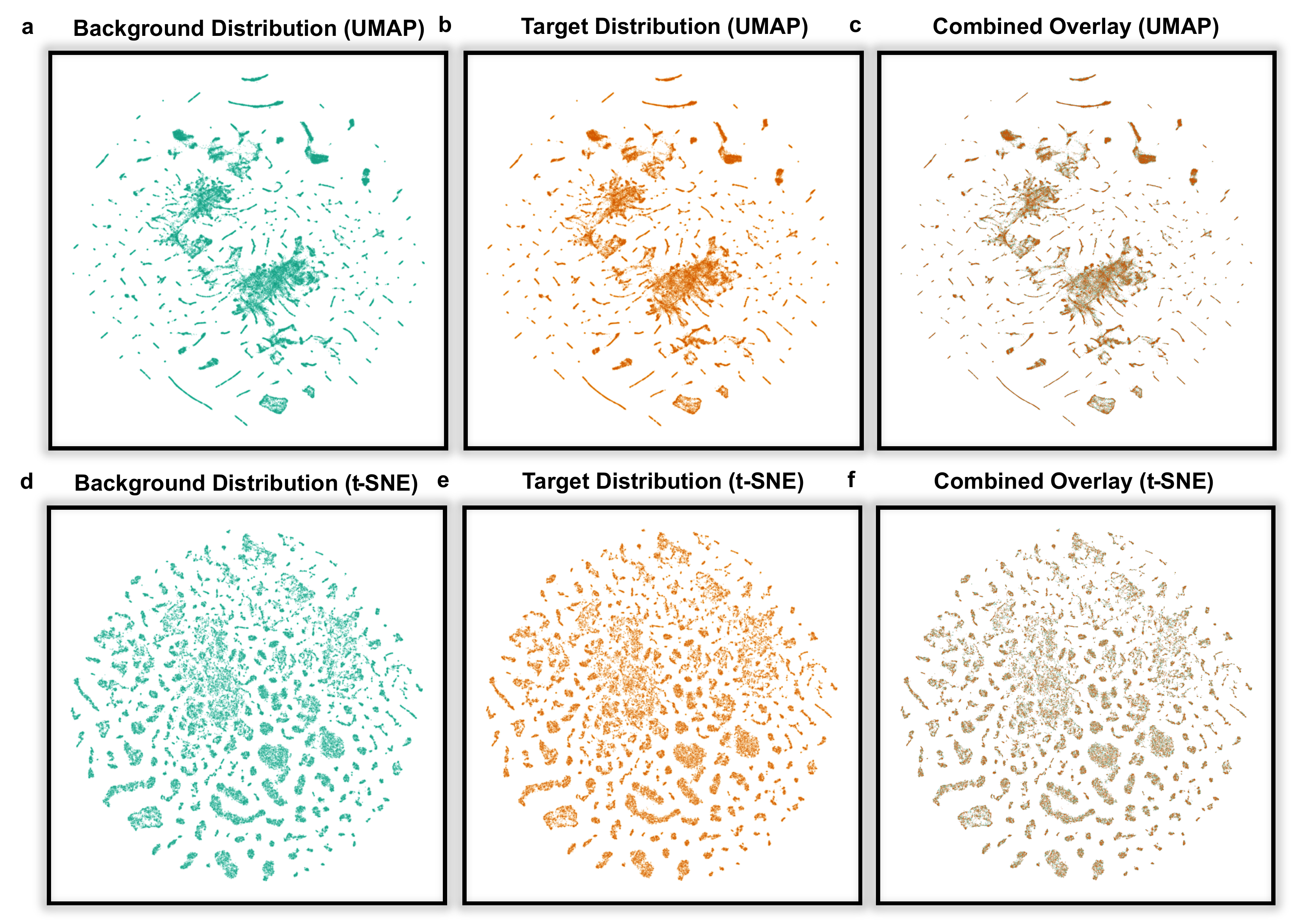}
	\caption{\textbf{Visualization of target and background feature distributions.} \textbf{a}, UMAP embedding of 130,000 randomly sampled pure background regions. \textbf{b}, UMAP embedding of 130,000 randomly sampled target bounding boxes. \textbf{c}, Combined overlay of both distributions. Because the target annotation is designed to capture the entire trajectory-constrained global morphology, background clutter inevitably occupies the dominant spatial proportion within these target boxes. Consequently, the feature distributions of the target boxes (\textbf{b}) and the pure background regions (\textbf{a}) are completely inseparable in the embedding space, as demonstrated by the perfect overlap in \textbf{c}. This quantitatively confirms that the visual features extracted from the target boxes are overwhelmingly governed by the surrounding environmental clutter rather than the target itself. Furthermore, the target samples are scattered across multiple distinct clusters rather than forming a compact group, verifying the extremely high intra-class variance of thermal wakes under diverse environmental conditions. \textbf{d–f}, Parallel analysis using t-SNE embeddings for background regions (\textbf{d}), target boxes (\textbf{e}), and their combined overlay (\textbf{f}). The consistent inseparability and high intra-class variance observed in both UMAP and t-SNE confirm that these phenomena are intrinsic to the data rather than artifacts of a specific dimensionality reduction technique.}
	\label{fig:umap}   
\end{figure*}

\section{Training Process Visualization}
\label{sec:Loss}
\begin{figure*}[!t]
	\centering
	\includegraphics[width=\linewidth]{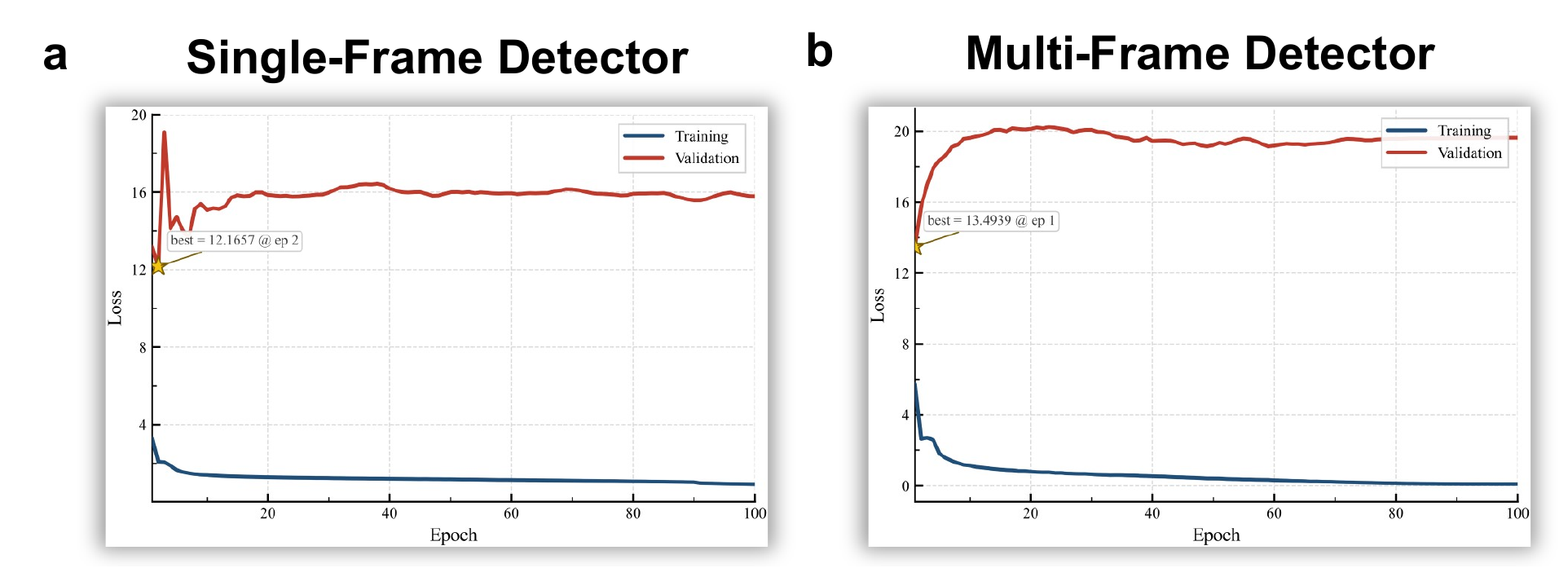}
	\caption{\textbf{Training process.} \textbf{a}, Training loss curve of Hyper-YOLO, a representative single-frame detector. \textbf{b}, Training loss curve of SSTNet, a representative multi-frame detector. Both detectors achieve near-zero AP on the test set, yet they converge successfully. However, their validation loss barely decreases, because changes in environmental conditions cause large simultaneous variations in both target and background characteristics. This highlights the stringent generalization requirement for detectors applied to unseen scenarios in real-world applications.}
	\label{fig:Loss}   
\end{figure*}

Many detectors achieve near-zero performance on the test set. Apart from the non-convergent cases we report, the remaining methods converge successfully. However, unlike many dataset splits that are independent and identically distributed, changes in external conditions induce substantial simultaneous shifts in both target and background characteristics. When a detector fails to learn an appropriate target representation, it cannot generalize even after convergence. Our split creates large differences in conditions between training and test sets, serving as an effective probe of whether a detector has learned suitable representations. Representative loss curves are shown in Figure \ref{fig:Loss}.

\section{Physical Basis of Trajectory--Morphology Consistency}
\label{sec:TrajMorph}
\noindent The trajectory--morphology consistency observed in the main text can be understood from the governing equation of the surface temperature field. Let $T(\mathbf{x},t)$ denote the temperature of the cold-skin layer on the free surface, $\mathbf{U}(\mathbf{x},t)$ the surface velocity field, and $\kappa$ the thermal diffusivity of water. The evolution of the surface temperature perturbation $\theta = T - T_0$ relative to the background field $T_0$ is governed by the advection--diffusion equation
\begin{equation}
	\frac{\partial \theta}{\partial t} + \mathbf{U}\cdot\nabla\theta = \kappa\nabla^2\theta + S(\mathbf{x}-\mathbf{r}_v(t)),
	\label{eq:theta}
\end{equation}
where $\mathbf{r}_v(t)$ is the instantaneous horizontal position of the vehicle along its trajectory, and $S$ is the source term representing the local thermal perturbation induced by volumetric displacement. The source term is spatially localized around the vehicle because the upward transport of subsurface water occurs in the immediate vicinity of the hull and the propeller wake.
Because the perturbation is weak, the nonlinear advection term can be approximated by a known surface flow. The formal solution of Eq.~\eqref{eq:theta} can then be written as a convolution with the Green's function $G$ of the advection--diffusion operator,
\begin{equation}
	\theta(\mathbf{x},t) = \int_{-\infty}^{t}\int_{\mathbb{R}^2}
	G(\mathbf{x}-\mathbf{x}',t-t')\,
	S(\mathbf{x}'-\mathbf{r}_v(t'))\,
	\mathrm{d}\mathbf{x}'\,\mathrm{d}t'.
	\label{eq:solution}
\end{equation}
Since the Green's function is spatially localized, the surface temperature anomaly at a given time is the superposition of contributions from the vehicle's past positions along its trajectory. In the limit where the thermal relaxation time is long compared with the characteristic time of the vehicle motion, Eq.~\eqref{eq:solution} reduces to
\begin{equation}
	\theta(\mathbf{x},t) \propto
	\int_{-\infty}^{t}
	S(\mathbf{x}-\mathbf{r}_v(t'))\,
	\mathrm{d}t'.
	\label{eq:integral}
\end{equation}
This integral is a line integral along the trajectory. Its spatial support therefore follows the vehicle's path, which explains why the overall wake morphology is self-similar to the trajectory, regardless of whether the path is straight or curved.

In contrast, the internal structure of the wake, whether smooth stripes or discrete vortical patches, is determined by the lateral structure of the source term $S$, not by the trajectory. The lateral scale of $S$ is set by local hydrodynamic instabilities such as vortex shedding and turbulent mixing, whose characteristic length is on the order of the vehicle diameter $D$ and whose evolution depends on depth, speed, and stratification rather than on the curvature of the path. These local instabilities are generated at random positions and times within the wake region, and their superposition produces the observed vortex patches. This separation of scales explains why the global morphology follows the trajectory while the internal texture remains trajectory-independent. When the vehicle turns, the line integral in Eq.~\eqref{eq:integral} curves accordingly, but the local source structure $S$ remains unchanged; when the vehicle hovers, the integral collapses to a point and the wake reduces to an isolated vortex.

\section{Geometric Basis of Camera-to-Wake Relative Positioning}
\label{sec:CameraWake}

\noindent The relative position between the camera and the wake can be approximated from the vehicle trajectory and the known camera pose. Let the horizontal position of the vehicle at time \(t\) be \(\mathbf{p}_v(t) = (x_v(t), y_v(t))\), and let the camera be located at \(\mathbf{p}_c = (x_c, y_c, h_c)\), where \(h_c\) is the camera altitude. Under a pinhole camera model, the image coordinates \(\mathbf{u} = (u, v)\) of a surface point \(\mathbf{p}_s = (x_s, y_s, 0)\) on the water surface are given by the projective transformation
\begin{equation}
	\begin{bmatrix} u \\ v \\ 1 \end{bmatrix}
	\simeq \mathbf{K} \, \mathbf{R} \,
	\begin{bmatrix} x_s - x_c \\ y_s - y_c \\ -h_c \end{bmatrix},
	\label{eq:projection}
\end{equation}
where \(\mathbf{K}\) is the intrinsic matrix and \(\mathbf{R}\) is the rotation matrix relating the camera frame to the world frame. If the wake remained directly beneath the vehicle, its image region would correspond to the projection of the vertical line through \(\mathbf{p}_v(t)\). This provides a first-order estimate of where to search for the wake in the image.
In practice, the wake can drift horizontally from the vertical projection of the vehicle. The drift is caused by surface currents, wind stress, and the residual motion of the cold-skin layer after the vehicle has passed. To first order, the horizontal displacement of the wake after a time \(\Delta t\) can be written as
\begin{equation}
	\Delta \mathbf{x}_w \simeq \mathbf{u}_s \, \Delta t,
	\label{eq:drift}
\end{equation}
where \(\mathbf{u}_s\) is the effective surface drift velocity and \(\Delta t\) is the time elapsed since the vehicle passed that location. Because \(\mathbf{u}_s\) is generally small and \(\Delta t\) is limited to the period during which the wake remains visible, the drift magnitude is typically modest. This explains why the relative position between the camera and the vehicle trajectory provides a useful but approximate cue for annotation.
In our workflow, this geometric cue is used only to restrict the search region for manual annotation. It is never treated as an exact target location, and it is always combined with trajectory consistency, hysteresis, CFD-assisted morphology prediction, and expert cross-validation. The purpose is to reduce the annotation search space while preserving the physical ambiguity of weak wakes.
\begin{figure*}[!t]
	\centering
	\includegraphics[width=\linewidth]{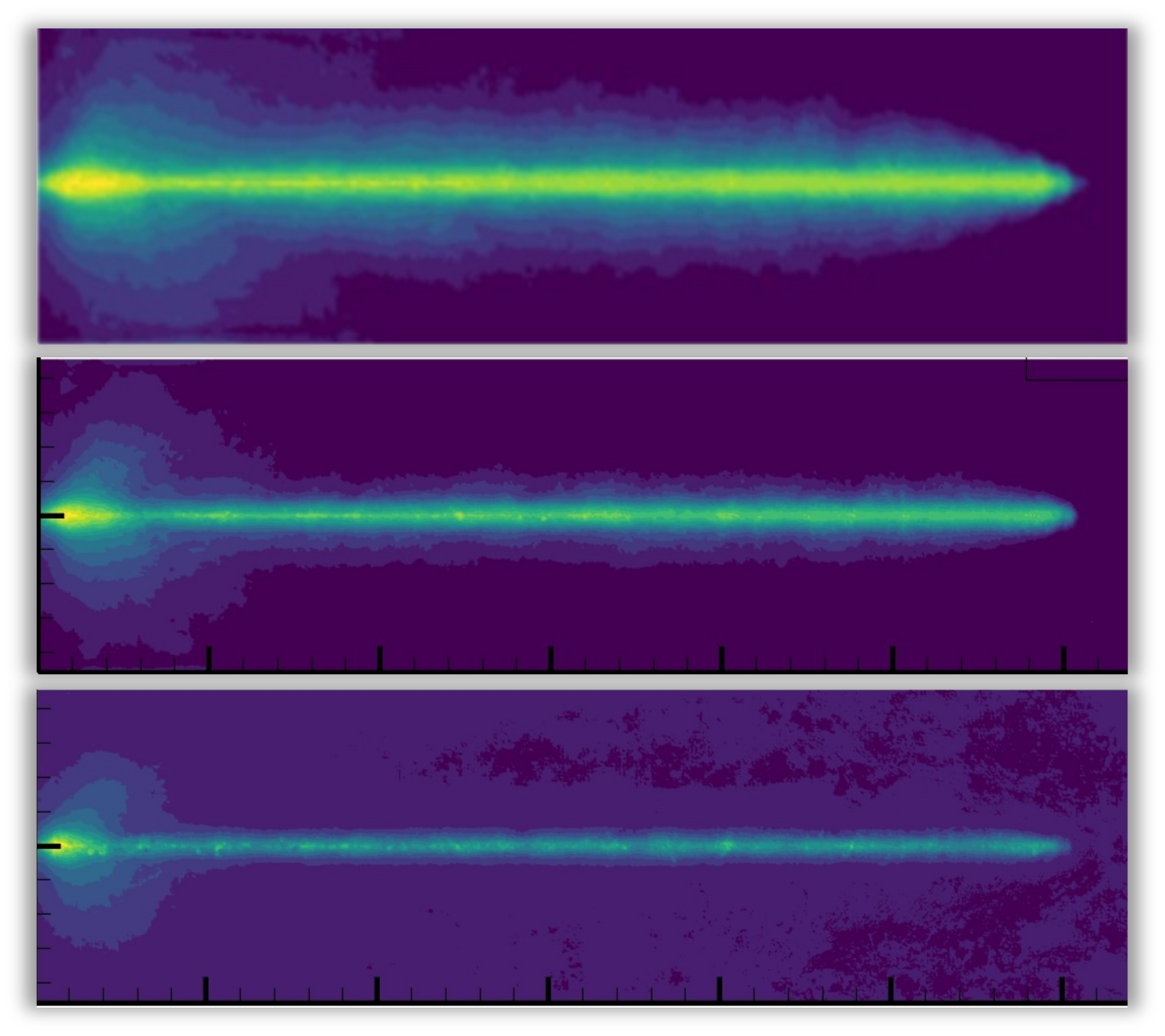}
	\caption{\textbf{CFD-simulated surface temperature fields under different operating conditions.}}
	\label{fig:CFD}   
\end{figure*}
\begin{figure*}[!]
	\centering
	\includegraphics[width=\linewidth]{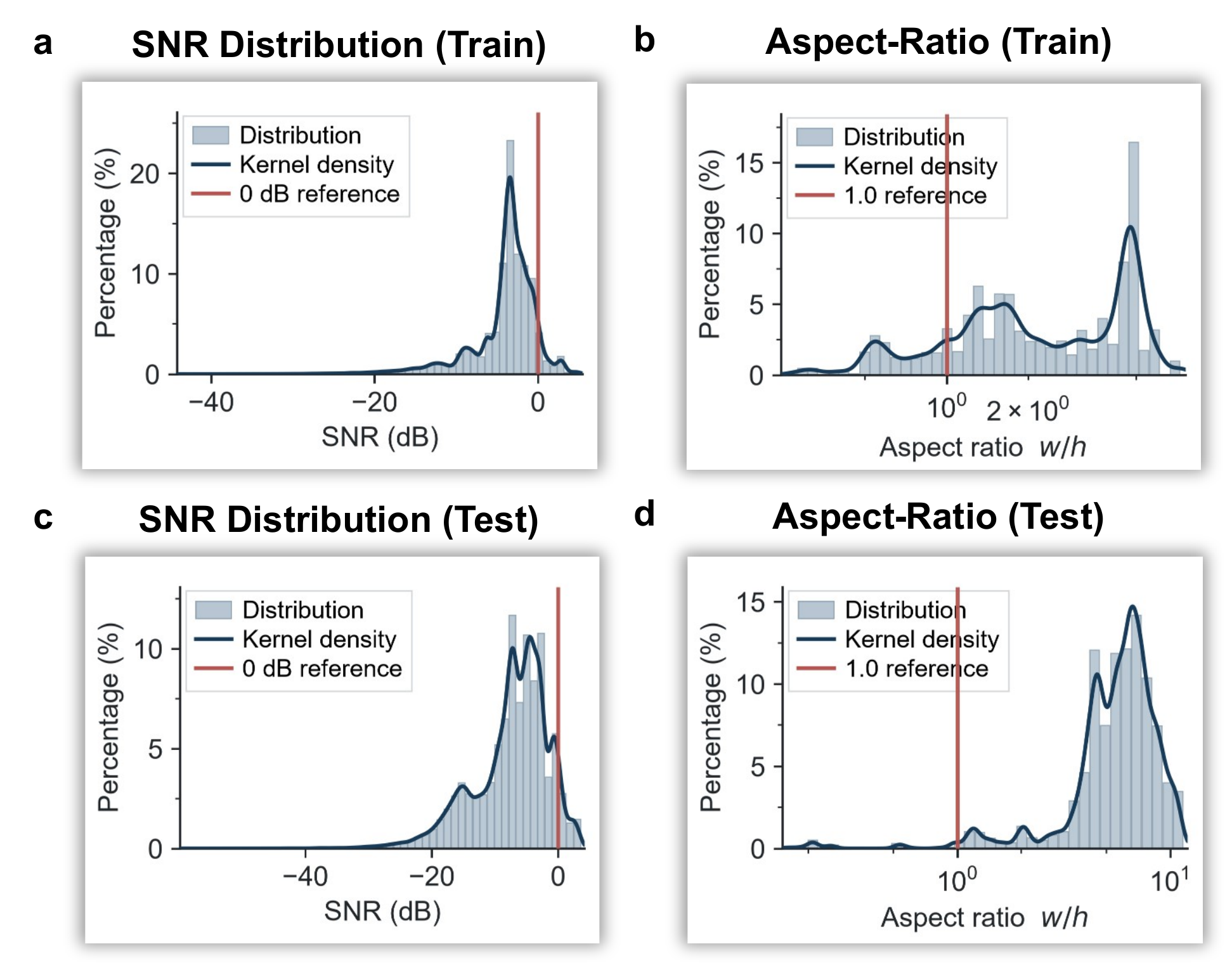}
	\caption{\textbf{Statistical characterization of AVP-IR-Wake.}}
	\label{fig:SDistribution}   
\end{figure*}

\section{Fréchet Inception Distance (FID) Analysis}
\label{sec:FID}
To quantitatively evaluate the distribution shift between the training and test sets, we compute the Fréchet Inception Distance (FID). Because our dataset is split at the acquisition-day level, the training and test sets contain entirely different environmental conditions (e.g., wind speed, solar radiation, and cold-skin state). Standard pixel-level comparisons fail to capture the structural and semantic differences induced by these environmental changes. FID, in contrast, measures the distance between the high-level feature distributions extracted by a pre-trained network, providing a robust metric for domain discrepancy.

We extract feature representations from all frames using a pre-trained Inception-v3 network. The feature vectors are then modeled as two multivariate Gaussian distributions: \(\mathcal{N}(\mu_t, \Sigma_t)\) for the training set and \(\mathcal{N}(\mu_v, \Sigma_v)\) for the test set. The FID is computed as the Fréchet distance between these two distributions:
\begin{equation}
	\mathrm{FID} = \|\mu_t - \mu_v\|^2 + \mathrm{Tr}\left(\Sigma_t + \Sigma_v - 2(\Sigma_t \Sigma_v)^{1/2}\right).
\end{equation}
The first term measures the distance between the feature means, while the second term captures the difference in feature covariances.
For our dataset, the computed FID is \textbf{252.467}. To contextualize this value, FID scores for typical independent and identically distributed (IID) splits of standard vision datasets are usually in the single or low double digits. A score of 252.467 signifies a profound distribution discrepancy. This large distance is not driven by sensor noise or annotation artifacts, but rather by the substantial simultaneous shifts in both target and background characteristics under different environmental conditions. For instance, a wake observed under calm nighttime conditions (deep, continuous stripe) and a wake observed under strong daytime winds (shallow, fragmented vortex clusters) map to vastly different regions in the feature space. This quantitative result directly corroborates the environment-dependent physical nature of thermal wakes, and underscores why detectors relying on local, environment-specific cues fail to generalize from the training set to the unseen test set.

\clearpage

\newpage

\renewcommand{\Crefname}{Supplementary References}

\bibliography{mybibfile}

\end{document}